\documentclass[11pt]{article}
\usepackage[a4paper,margin=1in]{geometry}
\usepackage[T1]{fontenc}
\usepackage[utf8]{inputenc}
\usepackage{textcomp}
\usepackage{lmodern}
\usepackage{graphicx}
\usepackage{amsmath,amssymb}
\usepackage{url}
\usepackage{hyperref}
\usepackage{caption}
\usepackage{booktabs}
\usepackage{float}
\usepackage{array}
\usepackage{longtable}
\usepackage{microtype}
\hypersetup{colorlinks=true,linkcolor=blue,citecolor=blue,urlcolor=blue}
\title{Game Theory Driven Multi-Agent Framework Mitigates Language Model Hallucination}
\author{Runzhe Liu$^{1,2}$, Biquan Bie$^{4}$, Zihao Wang$^{3}$, Yuchao Ma$^{1,2}$, Yexin Liu$^{3}$, Xinghai Li$^{1,2}$\\ Harry Yang$^{3}$, Wenbo Yang$^{1,2,*}$, Jinzhe Cao$^{1,2}$, Shengyang Tao$^{1,2,*}$}
\date{}
\begin{document}
\maketitle
\begin{center}
\small
$^{1}$State Key Laboratory of Fine Chemicals, Frontier Science Center for Smart Materials, Dalian University of Technology, Dalian, 116024, China\\
$^{2}$Dalian Key Laboratory of Intelligent Chemistry, CR Belt and Road Joint Laboratory on Intelligent Chemistry and Advanced Materials of Liaoning Province, School of Chemistry, Dalian University of Technology, Dalian, 116024, China\\
$^{3}$Academy of Interdisciplinary Studies, The Hong Kong University of Science and Technology, Clear Water Bay, Kowloon, Hong Kong\\
$^{4}$Independent Researcher, Beijing 100032, China.\\
$^{*}$Correspondence: \href{mailto:wbyang@dlut.edu.cn}{wbyang@dlut.edu.cn}; \href{mailto:taosy@dlut.edu.cn}{taosy@dlut.edu.cn}
\end{center}

\begin{abstract}
The application of lightweight Large Language Models in rule-based scientific domains remains severely limited by their tendency to mimic linguistic patterns rather than reproduce axiomatic reasoning, causing frequent hallucinations. Here, we show that G-Frame, an adaptive multi-agent framework integrating Bayesian and team game principles, establishes an automated closed-loop for high-quality data synthesis and model training. By forcing the internalization of domain constraints through structured reasoning, we synthesized a specialized corpus of 363,045 chains-of-thought and 199,589 question-answer pairs. The resulting 7B model OmniChem achieves performance parity with GPT 4o mini on custom benchmarks and ChemBench while exhibiting a 79.46\% reduction in hallucinations relative to its base architecture. We further demonstrate the advanced capabilities of OmniChem in molecular design and synthesis planning. This work establishes a scalable paradigm utilizing adaptive multi-agents to overcome inherent reasoning deficiencies, offering a feasible pathway for accelerating knowledge discovery in specialized scientific fields.
\end{abstract}

\section{Introduction}
Transformer-based Large Language Models (LLMs) show significant potential to transform scientific research in knowledge discovery and experimental design\cite{ref1,ref2,ref3,ref4,ref5,ref6,ref7,ref8,ref9,ref10}. Lightweight LLMs offer a compelling alternative by providing secure and scalable local deployment options versus closed-source commercial APIs\cite{ref11,ref12,ref13,ref14,ref15}. Task-specific fine-tuning allows these smaller models to achieve performance parity with larger counterparts at significantly reduced computational costs.

Applying general-purpose LLMs to rigorous scientific fields like chemistry faces fundamental challenges. Their probabilistic autoregressive nature struggles to replicate structured expert reasoning or internalize axiomatic physical constraints. This limitation results in plausible yet factually incorrect text generation\cite{ref1,ref16,ref17}. Such hallucinations are particularly acute in lightweight models and obstruct their reliable application in drug discovery or materials design\cite{ref18,ref19,ref20,ref21,ref22}.

Existing approaches including tool integration or supervised fine-tuning have inherent limitations\cite{ref23,ref24,ref25,ref26,ref27,ref28,ref29,ref30,ref31,ref32,ref33,ref34,ref35}. Reliance on closed-source APIs compromises data sovereignty while offline model fine-tuning is restricted by specialized data availability. This study therefore addresses the generation of factually unsupported text by lightweight LLMs within specialized domains like chemistry. This issue stems from the failure of autoregressive mechanisms to reproduce structured causal reasoning, and its resolution is critical for enhancing model reliability and reasoning in complex scientific tasks.

Mitigating factual inaccuracies arising from inadequate domain reasoning requires internalizing physical principles via complex training optimization. To address this challenge, we developed G-Frame. This adaptive multi-agent framework integrates game-theoretic principles to reframe data generation and model training. The system functions as a hierarchical probabilistic optimization architecture (Fig. 1). A cooperative Team Game operates at the micro-level to structurally suppress the entropy accumulation inherent in autoregressive generation. A Bayesian Game functions at the macro-level to dynamically optimize decision-making strategies under uncertainty. The framework establishes an automated closed-loop for data synthesis and adaptive training through this dual-layer dynamic policy iteration. This design explicitly enhances the domain reasoning of the model and internalizes physical principles.

\begin{figure}[!htbp]
\centering
\includegraphics[width=\linewidth]{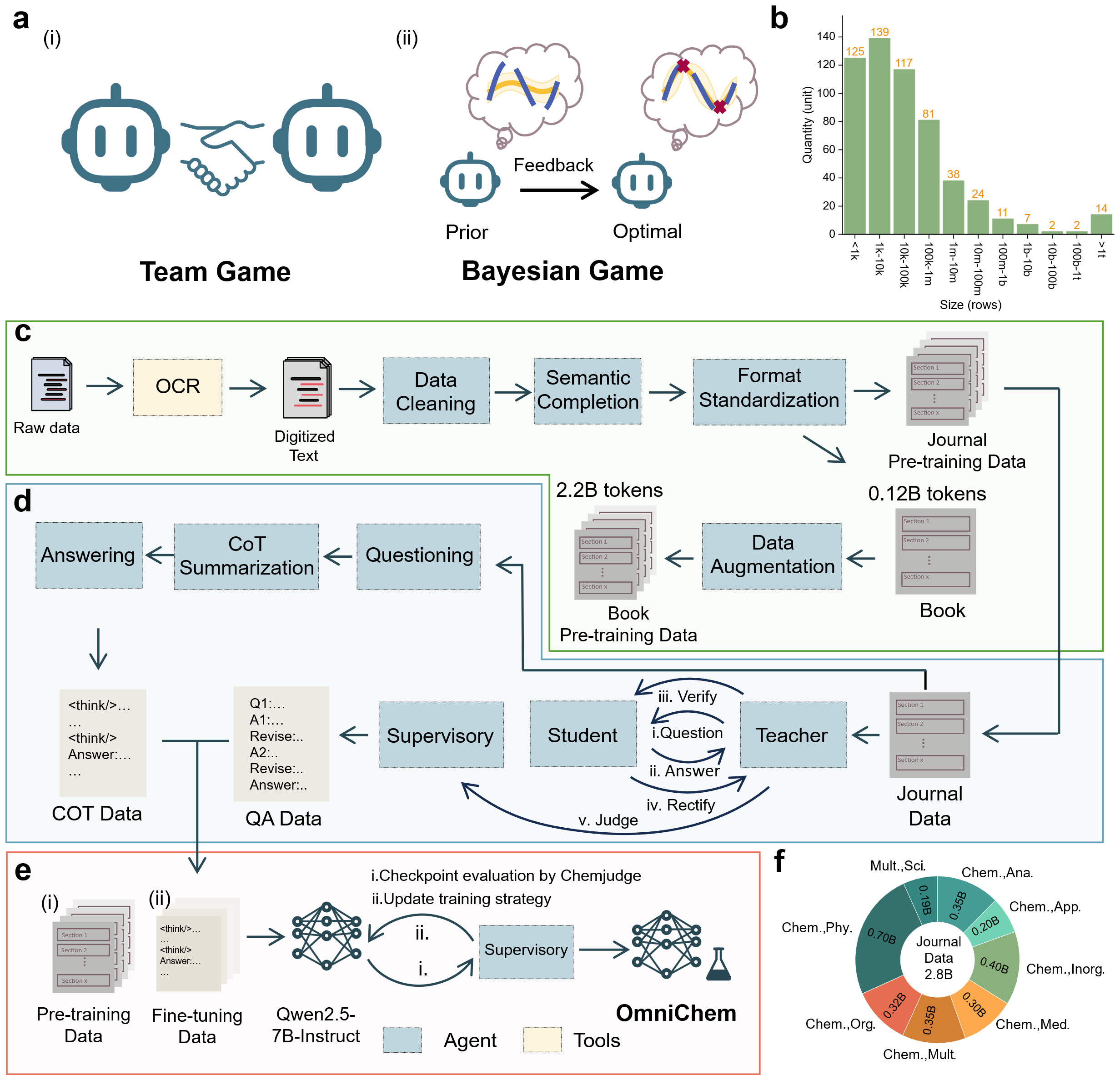}
\caption{a. G-Frame achieves adaptive strategies through two distinct modes: (i) team games and (ii) Bayesian games. b. The bar chart illustrates the number of open-source databases designated with chemical labels on Hugging Face as of May 15, 2025. Notably, a substantial portion of these datasets contains invalid entries or pertains to other domains; thus, the amount of genuinely usable data for chemistry is less than the reported. c, d, e. The workflow for building OmniChem using G-Frame is illustrated, consisting of three modules: data preprocessing, data synthesis, and model training. f. The composition of the journal corpus used for pre-training is shown.}
\label{fig:fig1}
\end{figure}

We utilized G-Frame to clean a 5-billion-token chemical corpus for continued pre-training and synthesized extensive chemical CoT and QA datasets to train OmniChem (Fig. 1c--f). The necessity of integrating domain pre-training, synthetic data fine-tuning, and adaptive training to mitigate catastrophic forgetting and minimize factual hallucinations is established by ablation analysis. This 7B model achieves performance on the ThChem and ChemBench\cite{ref36} benchmark comparable to GPT-4o mini and approaching GPT-o3. The ChemJudge test confirmed a significant reduction in factual inaccuracies relative to the base model. OmniChem further demonstrates advanced capabilities in autonomous report generation, knowledge graph construction and retrosynthesis planning. This framework establishes a scalable paradigm for applying artificial intelligence in science.

\section{G-Frame Game-Theoretic Multi-Agent Framework}
\begin{figure}[!htbp]
\centering
\includegraphics[width=\linewidth]{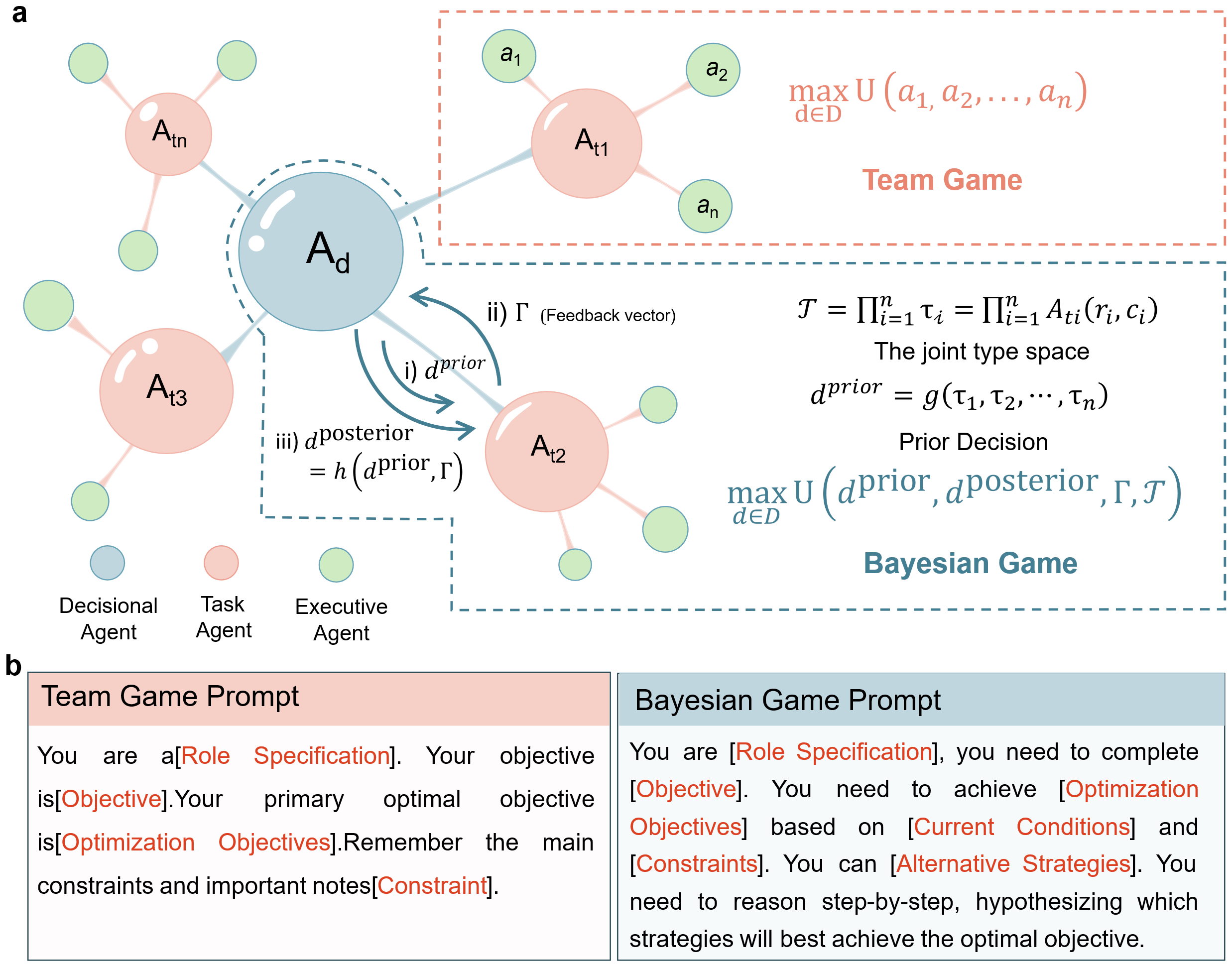}
\caption{Architecture of G-Frame and its team-game and Bayesian-game mechanisms. The framework couples worker-agent role constraints with a decisional agent that updates prior and posterior decisions from real-time feedback. Prompt templates for team and Bayesian games are shown in panel b.}
\label{fig:fig2}
\end{figure}

Inspired by team and Bayesian game principles for complex task resolution, the G-Frame adaptive multi-agent framework was developed to construct and optimize lightweight LLMs. Unlike single-model approaches, G-Frame employs multiple LLM agents within a game theoretic structure where collaboration is governed by interaction protocols and dynamic policy adjustments (Fig. 2a). This framework integrates data processing with adaptive training to enhance specialized domain capabilities while suppressing knowledge-based errors.

G-Frame employs a multi-agent collaborative model grounded in team game theory. This approach utilizes hierarchical task decomposition to circumvent the high entropy and reasoning failures inherent to lightweight LLMs. Limitations in network capacity often prevent lightweight models from maintaining coherent reasoning for complex multidimensional tasks. This deficiency causes probabilistic mass leakage into erroneous regions and manifests as hallucinations. To mitigate this G-Frame decomposes macroscopic tasks into series of well-defined subtasks each handled by a dynamically configured task agent. A task agent is itself a micro multi-agent system comprising two to three executive agents with granular roles. They collaborate by sharing intermediate states to maximize their pre-set utility functions. This process functions effectively as a distribution refinement mechanism. The construction of each utility function is made possible by precise prompt engineering, which defines the agent's task and optimization objective based on the LLM's operational principle of mapping inputs to a vector space (Fig. 2b). Decomposition into short subtasks restricts the information entropy at each step. This significantly enhances execution capability and factual accuracy in complex scenarios.

A second core mechanism addresses the fundamental shortcomings of LLMs as myopic autoregressive models for long-horizon decision-making under uncertainty. Their sequential token generation lacks global planning capabilities and a dynamic belief state for managing uncertainty, resulting in error accumulation and policy deviation in long-chain tasks. G-Frame implements an adaptive optimization mechanism grounded in Bayesian game principles. This mechanism addresses multidimensional control problems formally modeled as a Partially Observable Markov Decision Process. A decisional agent monitors the system state and formulates a prior policy on workflow and operational parameters. This prior is derived from expectations of future states within the joint type space of task agents. Continuous feedback is then used to iteratively update the agent's belief about this type space.. This yields an optimized posterior decision that maximizes a predefined global payoff function. This structured prior-posterior update cycle compels the LLM to operate as a constrained policy proposer rather than an open-ended generator. It establishes an adaptive closed loop that compensates for model deficiencies in uncertainty quantification.

Fundamentally G-Frame functions as a hierarchical probabilistic optimization system. We formally model the micro-level collaboration as a distribution refinement mechanism designed to structurally minimize the entropy of generated text. This suppresses hallucinations by minimizing the KL divergence between the generated and true distributions. Concurrently macro-level decision-making is formalized as a solver for the Partially Observable Markov Decision Process. This maximizes global utility under strictly defined physical constraints (Key mathematical definitions are provided in the Supplementary Information).

The G-Frame adaptive multi-agent framework fuses a team-game collaborative mechanism with a Bayesian adaptive decision-making mechanism. This three-layer architecture comprises decisional, task, and executive agents. Within this structure, task agents are defined for data cleaning, data augmentation, synthetic data generation, and adaptive training, each employing distinct game strategies.

The quality of pre-training corpora is critical, as noise from Optical Character Recognition of chemistry texts introduces knowledge deficiencies that cause model inaccuracies. G-Frame's multi-agent, team-game strategy addresses this challenge. A task agent directs three executive agents responsible for symbol cleaning, semantic completion, and structural normalization . Agent collaboration toward a common objective, a character error rate below 0.5\% and punctuation accuracy above 99\% systematically improves accuracy and overcomes the deficiencies of single-model text processing .

To address the limited data scale and insufficient stylistic coverage of book texts in the pre-training stage, a Data Augmentation Agent was designed within G-Frame. This agent employs a team game-based collaborative model utilizing six distinct augmentation strategies to generate text variants with diverse narrative styles by simulating novice, intermediate and advanced expert personas .

For the supervised fine-tuning stage, a large volume of CoT and QA data was synthesized. The generation of high-quality QA data is crucial for enhancing the model's instruction-following and complex reasoning abilities, which ultimately suppresses hallucinations. A dedicated QA task agent operating within a team game model manages data generation by deploying three executive agents simulating Teacher, Student and Supervisor roles. A multi-round generate-evaluate-correct cycle is used to enhance the quality and logical rigor of each QA pair, with the Regulator agent providing final arbitration. The effectiveness of this process was validated on the SQuAD benchmark, where the multi-agent approach from G-Frame increased the F1 score by 20\%-40\% across lightweight models of various scales when compared to a single LLM (Fig. 3a).

To internalize the structured causal reasoning logic specific to the chemistry domain and thereby suppress hallucinations arising from reasoning deficiencies, a chemical CoT corpus was synthesized via data distillation and a CoT multi-agent approach. The model was subsequently fine-tuned with this corpus (detailed descriptions are provided in the Methods).

Engineering optimization of the G-Frame data processing workflow was performed to address the disparate computational demands of data synthesis tasks. A central feature is the adaptive control of asynchronous concurrency, a mechanism governed by the decisional agent that integrates Bayesian game theory. An initial concurrency level is established as a prior decision based on hardware and engine parameters. The agent updates this prior to a posterior decision by dynamically adjusting concurrency based on real-time operational feedback covering active processes request queue length and KV cache occupation. The efficacy of this adaptive mechanism was validated on a high-throughput CoT synthesis task, where service interruptions were prevented. Processing throughput was improved by approximately 22-fold relative to synchronous mode and by 25\% and 6.5\% relative to fixed concurrencies of 50 and 100, respectively (Fig. 3b).

To ensure the effective acquisition of domain knowledge for hallucination mitigation, G-Frame's adaptive principles were further applied to the entire model training process. The inherent uncertainty in optimizing training hyperparameters is managed through a Bayesian game mechanism, which enables the decisional agent to control the process autonomously. A prior policy on hyperparameter settings is formed based on initial data characteristics and hardware status. This policy is then updated to an optimized posterior via a Bayesian update mechanism, which is guided by payoff signals from the ChemJudge evaluation method. ChemJudge leverages an LLM referee to conduct multidimensional checkpoint performance scoring spanning hallucination accuracy and output structure. This posterior policy can then automatically trigger operations such as dynamic learning rate adjustment or selective data augmentation. Comparative pre-training experiments focusing on the Diels-Alder reaction (Fig. 3) demonstrate that while the unoptimized base model exhibits fundamental errors including incorrect reaction definitions and thermodynamic conceptual confusion a fixed-strategy model effectively resolves the core problem. The adaptively trained OmniChem model conversely generates a chemically accurate explanation identifying the concerted mechanism and detailing solvent stereoselectivity effects (Fig. 3d) to demonstrate robust physical rule internalization.

The independent mechanisms and synergistic dependencies of the three core training stages are elucidated by the ablation study. A deficit in specialized chemical knowledge is observed when domain pretraining is omitted. Repetitive generation is mitigated and instruction following capabilities are enhanced by synthetic data fine tuning. A distinct catastrophic forgetting dilemma is inherently introduced by this full parameter update process. The resulting performance decline is rectified and boundary behavior is aligned through adaptive training to establish a balance between factual reasoning and generalization capabilities (Detailed discussion and the full ablation table are provided in the Supplementary Information).

\section{G-Frame for OmniChem}
The adaptive characteristics of G-Frame's game theoretic mechanisms were demonstrated in the chemistry domain. A 5-billion-token pre-training corpus was yielded from the processing of approximately 500,000 chemistry articles and books. Subsequently, approximately 363,045 chemical CoT and 199,589 QA pairs were synthesized. This data enabled the adaptive supervised training of OmniChem, a 7B parameter chemistry reasoning LLM.

The performance of OmniChem was evaluated using the ChemJudge, ThChem and ChemBench benchmarks. On ChemJudge a 471-question test judged by Gemini 3.1 Pro a hallucination scores superior to GPT 4o mini and a 79.46\% reduction in hallucination rate relative to the base model were observed . On the ThChem chemistry reasoning benchmark and ChemBench accuracy rates of 79.45\% on ThChem 1.0 and 62.08\% on ThChem 2.0 and 49.82\% on ChemBench were achieved as illustrated in Figure 4a. (detailed design is described in the Methods). The performance of the 7B OmniChem model was thereby shown to be comparable to GPT 4o mini reaching 80\% to 97\% of the GPT o3 level.

\begin{figure}[!htbp]
\centering
\includegraphics[width=\linewidth]{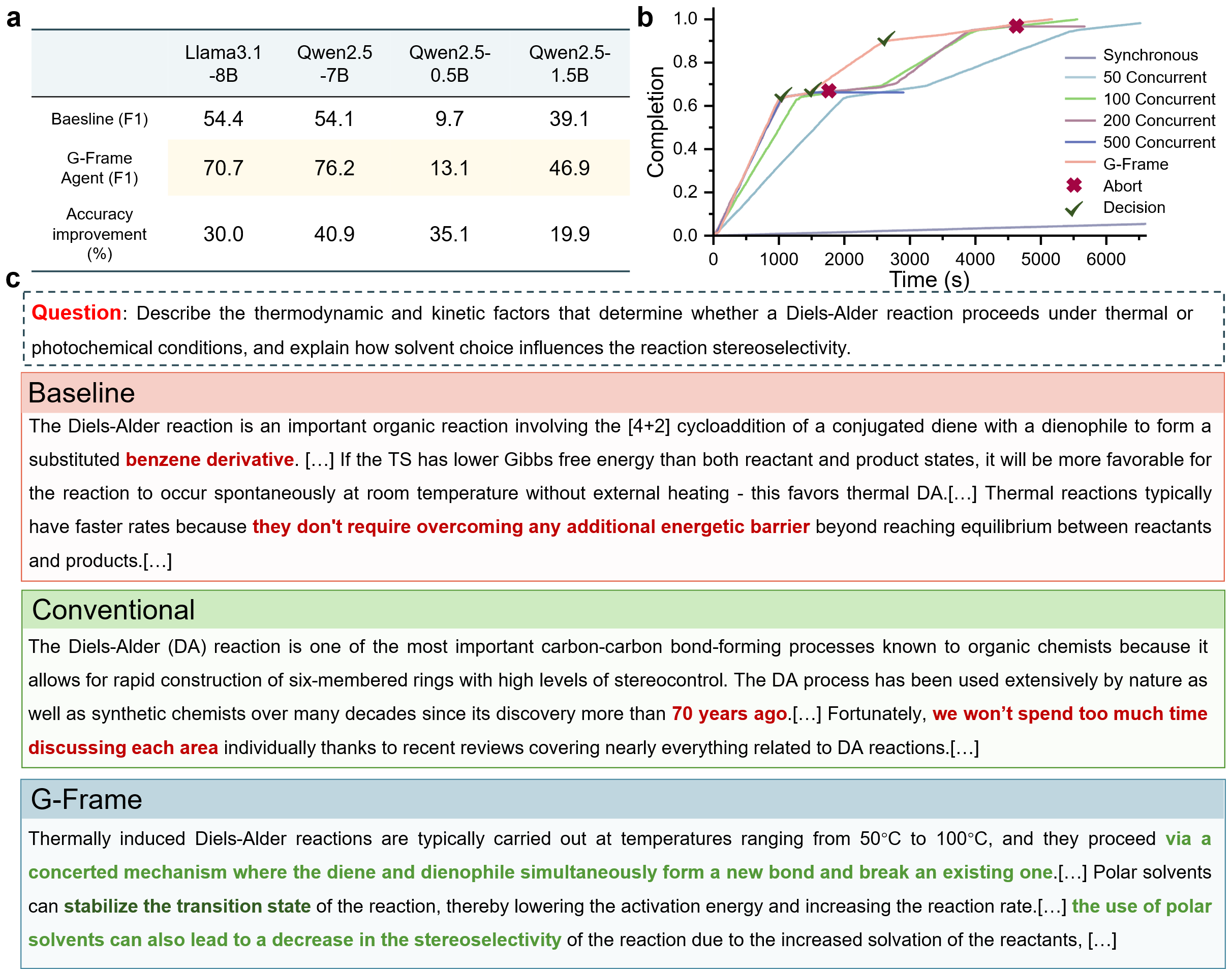}
\caption{a. The performance of G-Frame compared to a single LLM on the SQuAD 2.0 benchmark is shown. b. The efficacy of G-Frame in concurrency optimization is demonstrated. c. A qualitative comparison of the question-answering performance for the three models pre-training.}
\label{fig:fig3}
\end{figure}

A multi-agent framework with a Deep Research function was constructed by OmniChem through the integration of academic search APIs, enabling the generation of extensive academic reports (Fig. 4b). This function was utilized to study "How to design efficient, stable, and color-pure deep-blue TADF materials", yielding a report that achieved scores reaching 90\% of those from GPT-o3 and Gemini in LLM-as-a-Judge evaluations (Fig. 4c) (the quantitative Deep Research evaluation is provided in the Supplementary Information). A QA Expert was also designed through the integration of GraphRAG, which supports the construction of knowledge graphs from domain-specific data. The subsequent use of retrieval-augmented generation techniques enhances answer accuracy and contextual relevance, thereby effectively reducing hallucinations when the lightweight model processes complex information.

We validated OmniChem's practical utility by benchmarking its performance across photophysical modulation physicochemical optimization and synthesis planning. The evaluation focused not on task-specific fine-tuning but on the model's ability to generate innovative design strategies. The model's chemical insight was demonstrated when tasked with designing a BODIPY derivative for deep-red light absorption, a valuable property for bio-imaging. The core principle of extending the \ensuremath{\pi}-conjugated system to reduce the HOMO-LUMO gap and thus red-shift the absorption wavelength was accurately applied. A D-\ensuremath{\pi}-A structure was proposed, where ethynylphenyl \ensuremath{\pi}-bridges at the 3- and 5-positions of the BODIPY core connect to an N,N-diphenylamino electron-donating group . DFT calculations validated this design prioritizing efficient intramolecular charge transfer for wavelength extension yielding a predicted deep-red maximum of 709 nm (Fig. 4d). The ability to optimize physicochemical properties was also shown by addressing the poor water solubility of the BODIPY core. Introducing arginine or sulfobetaine moieties at the 8-position leverages corresponding hydrogen bonding and ion-dipole interactions to establish stable hydration shells. This substantial enhancement in water solubility, critical for biological applications, was quantified by a calculated decrease in logP from 1.82 to 1.11 for arginine and -4.44 for sulfobetaine (Fig. 4e). Finally, the model's application in retrosynthetic planning was evaluated using the anesthetic lidocaine. A concise two-step route was planned, involving the acylation of 2,6-dimethylaniline with chloroacetyl chloride, followed by nucleophilic substitution of the chloroamide intermediate with diethylamine. The consistency of this route with the established commercial synthesis confirmed the model's capacity for generating chemically sound and practical schemes (Fig. 4f).

\begin{figure}[!htbp]
\centering
\includegraphics[width=\linewidth]{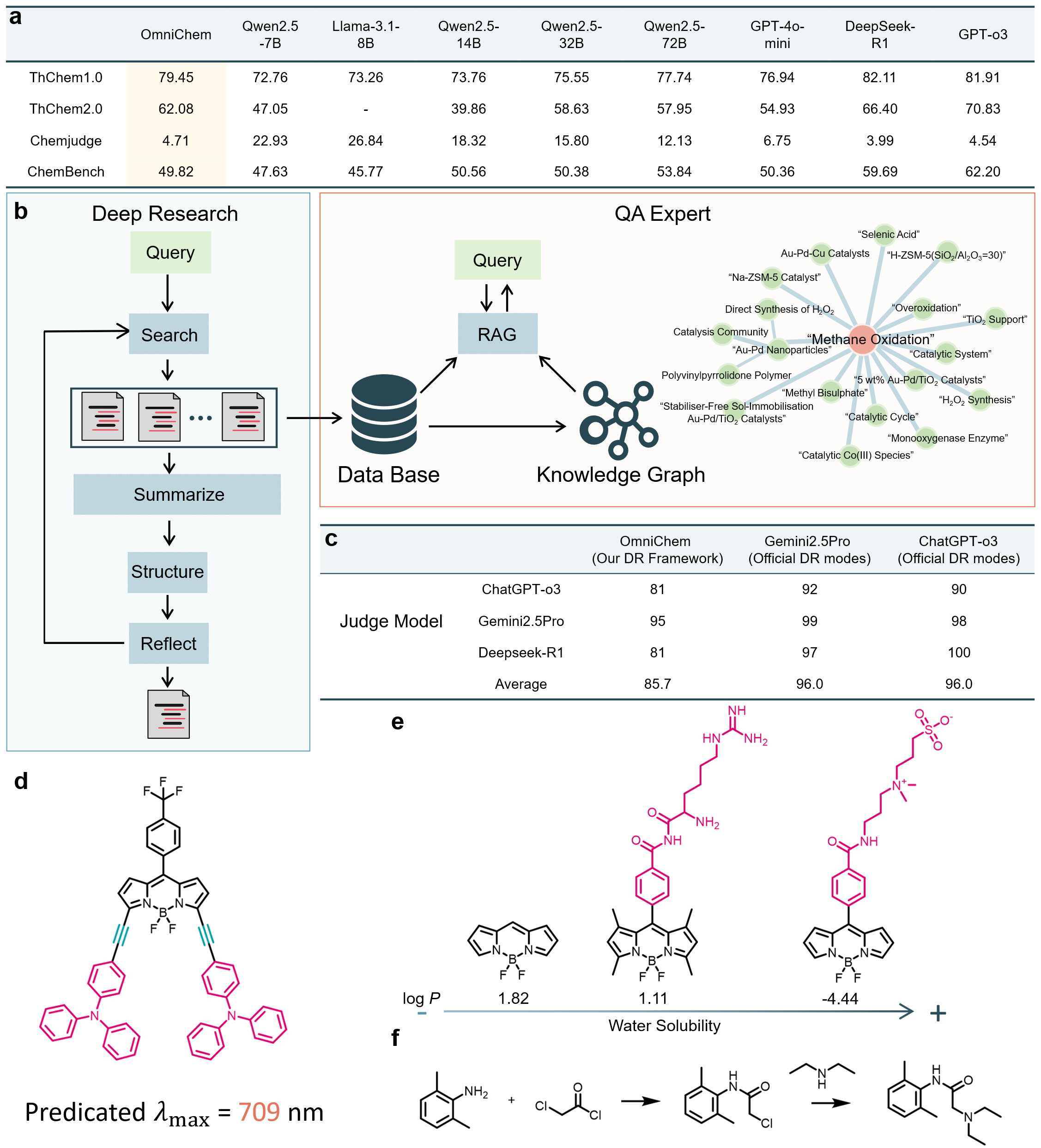}
\caption{a. Scores of different models on the ThChem1.0, ThChem2.0, and ChemJudge benchmarks. The ThChem benchmarks use a 100-point scale, whereas ChemJudge is evaluated based on the hallucination rate. A dash (-) indicates that the model does not apply to this test. b. Flowchart illustrating the workflow for a deep researcher and domain knowledge QA Expert, with a visually enhanced knowledge graph shown as an example. The original graph is provided in the Supplementary Information. c. Comparative evaluation of deep research reports. Quantitative assessment of reports generated by OmniChem (Our DeepResearch Framework) versus Gemini 2.5 Pro and ChatGPT-o3 (Official Deep Research modes). The evaluation employs an LLM-as-a-Judge protocol, assigning scores across four distinct dimensions (25 points each, totaling 100): Information Accuracy, Depth and Logic, Structure and Expression, and Novelty and Utility. d. Demonstration of OmniChem's chemical insight when tasked with designing a BODIPY derivative for deep red-light absorption. e. Enhancing the water solubility of the BODIPY core structure using OmniChem. f. Synthesis planning for the small molecule drug lidocaine using OmniChem.}
\label{fig:fig4}
\end{figure}

The preceding examples demonstrate the capability of the G-Frame-trained OmniChem to address real-world research problems. This effectiveness is attributed to a training paradigm designed to mitigate the inherent limitations of lightweight LLMs systematically. The adaptive pre-training stage internalizes physical constraints of the chemistry domain to construct a knowledge representation consistent with the real world, addressing the lack of a physical anchor. Subsequent CoT fine-tuning encodes the structured causal reasoning patterns of experts, resolving the model's inability to reproduce rigorous logical chains. This training paradigm, which provides dual physical and logical constraints, is designed to effectively mitigate the inherent reasoning deficiencies of LLMs and provide reliable AI assistance for scientific research.

\section{Discussion}
In summary, the G-Frame adaptive multi-agent framework was constructed to address the hallucination problem in lightweight domain-specific models, which arises from an inability to support structured logical reasoning and internalize axiomatic physical constraints. The framework integrates Bayesian and team game theory to establish an automated, closed-loop process from high-quality data synthesis to adaptive model training, which internalizes these physical constraints and reasoning patterns. The framework's efficacy was validated through the construction of OmniChem, a 7B-parameter chemistry reasoning model whose excellent performance in molecular design and reaction planning was demonstrated. To promote further research, the resulting question-answering and CoT datasets will be made openly available. This research establishes a scalable, multi-agent development paradigm for constructing lightweight LLMs in specialized domains, offering a systematic solution applicable to other fields.

\section{Methods}
\subsection*{The G-Frame Framework}
The asynchronous communication and collaboration within the G-Frame multi-agent framework are implemented using the AsyncOpenAI library. In this system, all local Large Language Models (LLMs) intended for inference tasks are efficiently encapsulated as local API services via the vLLM library. The system also supports the construction of the G-Frame framework using APIs from commercial LLMs. Unless otherwise specified, the system defaults to using the Qwen2.5-7B-Instruct model as the core local LLM, with YaRN mode enabled to optimize long-context processing capabilities. The computational hardware environment primarily consists of NVIDIA A100 80GB or NVIDIA H100 80GB GPUs.

Within the G-Frame framework, we have integrated the DeepSeek-R1 API service provided by Volcano Engine for the decisional agent. Concurrently, to accommodate the varying hardware conditions of different researchers, G-Frame also supports the encapsulation of locally deployed LLMs as API services for integration into the framework.

During the system initialization phase, the decision-making agent engages in a multi-turn interactive dialog with the user to accurately capture their requirements. Based on this, it dynamically plans a complete and customized workflow for solving the scientific problem. The automated execution process is initiated only after G-Frame receives explicit confirmation of the workflow from the user. The execution of the entire workflow is scheduled and controlled by G-Frame through programmatic calls to the operating system's command-line interface, enabling the automated operation of all program modules. Each task agent within the framework is designed and encapsulated as an independent Python application, which can be invoked via the command-line interface and supports flexible parameter configuration to adapt to different task requirements.

\subsection*{Decoupled Asynchronous Communication Architecture}
The framework's real-time adaptive control is enabled by a decoupled, asynchronous multi-queue communication architecture, which adheres to a Multi-Producer, Single-Consumer event-driven pattern. The computational engine and task execution layers function as two independent producers, asynchronously placing structured data objects detailing the server's physical state (DataPoint\_V) and macroscopic task progress (DataPoint\_T) into their respective dedicated asyncio queues. Queue instances. The central decision agent acts as the sole consumer, utilizing concurrency primitives to listen for data events from both queues simultaneously. To ensure temporal consistency, the agent implements a synchronization protocol based on a shared timestep. Upon receiving a matching pair of heterogeneous data points, it merges them into a unified, high-dimensional System State Vector, which provides a complete and informationally consistent snapshot of the system's state. This mechanism guarantees that the decision agent receives high-quality, real-time feedback, providing a solid data foundation for its subsequent decision-making process.

\subsection*{Pseudocode for Team and Bayesian Games}
Here we present the pseudocode for Team games and Bayesian games.

\begin{figure}[!htbp]
\centering
\begin{minipage}{0.96\linewidth}
\small
\begin{verbatim}
Input: M: A macroscopic complex task. Din: Initial input data for
    task M. R= (r1, r2,... ,rn): An ordered sequence of n pre-
    defined roles for Execution Agents. U (a1, a2,... ,an): The
    global utility function that the entire process is designed to
    maximize. Output: Dout: The final processed output.
 1  Function Sequential Team Game (M, Din, R, U)
 2  // Step 1: Initialization
 3  Task Agent <- Instantiate Task Agent(M)
 4  Aseq= (A1, A2,... ,An) Task Agent. Instantiate Agents in
    Sequence(R)
 5  Current Data <- Din
 6  // Step 2: Sequential Execution Pipeline
 7  for each Execution Agent Ai in Aseq do
 8  // The goal of each agent is directly defined by its prompt
 9  Prompti <- Create Prompt (Ai. role, Current Data)
10  // Each agent is essentially an LLM call
11  LLM Outputi <- Ai. Call LLM(Prompti)
12  // The output of one agent becomes the input for the next
13  Current Data <- LLM Outputi
14  end for
15  // Step 3: Final Output Generation
16  Dout <- Current Data
17  return Dout
18  End Function
\end{verbatim}
\end{minipage}
\caption{Pseudocode for the G-Frame team game.}
\label{alg:team-game}
\end{figure}

\begin{figure}[!htbp]
\centering
\begin{minipage}{0.96\linewidth}
\small
\begin{verbatim}
Input: Sprior: Prior knowledge of the system state. T: Joint type
    space of Task Agents. Uglobal: The global utility function to be
    maximized. Output: A sequence of optimized decisions applied
    throughout the task lifecycle.
 1  Function Dynamic Bayesian Adaptive Decision(Sprior, T, Uglobal)
 2  // Step 1: Initialization
 3  Decisional Agent <- Get Decisional Agent ()
 4  dcurrent <- Decisional Agent. Formulate Initial Decision(Sprior,
    T)
 5  Apply Decision to System(dcurrent)
 6  // Step 2: Start the dynamic adaptation loop
 6  while Main Task Is Not Finished() do
 7  // The previous posterior decision becomes the current prior
 8  dprior <-dcurrent
 9  // Receive new information
10  Gamma <- Get_Real_Time_Feedback()
11  // Formulate posterior decision by maximizing the global utility
    function
12  // This step corresponds to dposterior = h(dprior, Gamma)
13  dposterior <-
14  Apply Decision to System(dposterior)
15  // Update the current state for the next iteration
16  dcurrent <- dposterior
17  end while
18  return "Process Complete"
19  End Function
\end{verbatim}
\end{minipage}
\caption{Pseudocode for the G-Frame Bayesian game.}
\label{alg:bayesian-game}
\end{figure}

\subsection*{Construction and Processing of Pre-training Data}
The model's pre-training dataset is composed of two components: domain-specific data and general-domain data. The domain-specific data includes over 500,000 chemistry-related scientific papers and approximately 1,000 core chemistry textbooks and monographs. Recognizing that a portion of the source literature is stored in PDF format with substantial image-based content, such as scanned pages, we employed the MinerU toolkit for high-quality Optical Character Recognition (OCR). Within the OCR workflow, document layout analysis was performed using the doclayout\_yolo model, while text region detection and recognition utilized the yolo\_v8 model with default language settings. Following OCR processing, the PDF documents were efficiently converted into a structured Markdown format. Subsequently, we used the Qwen2.5-7B-Instruct model for in-depth processing and fine-grained cleaning of these Markdown files to construct a high-quality, chemistry-focused pre-training corpus.

To mitigate the issue of catastrophic forgetting, which can occur when training a model exclusively on specialized data, and to maintain its general knowledge capabilities, we strategically incorporated general-domain corpora into the pre-training data. This portion of the data was primarily sourced from the widely used C4 and Wikipedia datasets. Finally, the chemistry-specific and general-domain corpora were mixed at an initial sampling ratio of 8:2 to form the input data for model pre-training.

\subsection*{Model Pre-training Process}
The model pre-training was conducted using the DeepSpeed distributed training framework, with AdamW employed as the optimizer and an initial learning rate of 3 \ensuremath{\times} 10\textsuperscript{-}\textsuperscript{5}. The process was structured into three training rounds. The first round completed a full 30,000 steps. The second round was concluded at approximately 8,000 steps via early stopping, which was triggered based on predefined performance metrics (e.g., loss function trends, validation set perplexity). For the third round, the optimizer's state was reset, and the sampling ratio of general-domain to chemistry-specific corpora was adjusted to 4:6. Throughout all rounds, a checkpoint was saved every 1,000 steps.

\subsection*{Model Supervised Fine-tuning Process}
Subsequently, full-parameter supervised fine-tuning of the model was performed using the DeepSpeed framework. The AdamW optimizer was used with an initial learning rate of 5 \ensuremath{\times} 10\textsuperscript{-}\textsuperscript{5}. This process consisted of two distinct training stages, each utilizing a different mixed dataset composed of QA and CoT data. Each stage lasted 24,000 steps, with checkpoints saved every 1,000 steps. The entire fine-tuning process was completed without the use of early stopping.

\subsection*{Construction and Processing of Fine-tuning Data}
The construction and processing of the model's fine-tuning data were dedicated to generating two critical types of data: CoT fine-tuning data and QA fine-tuning data. The core objective of the CoT data is to enable the model to learn and internalize the detailed reasoning paths and rigorous logical steps characteristic of human problem-solving in complex chemistry scenarios. To this end, we employed two complementary strategies. The first is knowledge distillation, utilizing a more capable "teacher model", DeepSeek-R1-Distill-Qwen-32B, to generate nuanced thought processes for specific chemical problems. The second strategy involves a three-LLM workflow: the first LLM poses a guiding question for a specific chemical problem; the second LLM generates a structured, summary Chain-of-Thought based on the question and relevant background knowledge; the third LLM then provides the final, professional answer based on this CoT. Both methods are designed to produce high-quality CoT data containing explicit thinking markers (e.g., intermediate steps enclosed in \textless{}think\textgreater{}...\textless{}/think\textgreater{} tags). The model temperature during this process was set to 0.6 (model Qwen2.5-14B-Instruct). Concurrently, the construction of QA data aims to train the model to generate answers that are well-structured, accurate, and conform to professional expression standards. This data was generated using the aforementioned framework, with the model temperature also set to 0.6. Crucially, throughout the entire generation process for both data types, to maximally mitigate model hallucination arising from incomplete information or contextual deficits, we consistently supplied the full text or core sections of relevant chemical literature as background knowledge to the LLM. This ensures that the model performs precise reasoning and provides reliable answers based on a comprehensive understanding of the problem's context.

\subsection*{F1 Score Calculation}
The F1 score is the harmonic mean of Precision and Recall. To understand the calculation, the following terms must first be defined based on the model's predictions:

True Positives (TP): The number of positive instances that were correctly classified as positive. False Positives (FP): The number of negative instances that were incorrectly classified as positive. False Negatives (FN): The number of positive instances that were incorrectly classified as negative.

1. Precision measures the proportion of true positives among all instances classified as positive. It quantifies the accuracy of the positive predictions.

2. Recall measures the proportion of true positives that were correctly identified from all actual positive instances. It quantifies the model's ability to find all positive samples.

3. The F1 score is a single metric that balances both Precision and Recall. As the harmonic mean, it gives more weight to lower values, meaning a high F1 score requires both high precision and high recall.

\subsection*{ThChem Benchmark}
ThChem is a benchmark meticulously constructed to accurately assess the comprehensive capabilities of large language models in the domain of chemistry. Its content extensively covers core branches of the discipline, including inorganic, organic, analytical, and physical chemistry. The benchmark comprises two iterative versions: ThChem1.0 and ThChem2.0. ThChem1.0 consists of 253 high-quality, multiple-choice questions. These questions not only assess fundamental knowledge---such as chemical equation calculations, substance structure inference, and experimental and characterization techniques---but also extend to advanced topics, including cutting-edge concepts in new material design and their potential applications. All questions are derived from real research cases and complex problems found in high-impact, peer-reviewed academic journals. They have been rigorously cross-validated by multiple senior chemistry experts to ensure scientific accuracy, clarity, and the uniqueness and correctness of the answers. Building upon this foundation, ThChem2.0 introduces a targeted modification. For approximately half of the questions randomly selected from ThChem1.0, the correct answer option was deliberately removed, rendering the standard answer as "null" or non-existent among the choices. This design is intended to evaluate the model's core abilities at a deeper level---specifically, its capacity for complex mechanism inference, multi-step logical reasoning, and the integrated application of chemical knowledge---rather than simply its ability to identify the correct answer from a given set of options.

\subsection*{ChemJudge Benchmark}
This benchmark utilizes the "LLM as a Judge" paradigm. It is composed of 471 in-depth chemistry questions, each designed to probe fundamental theories, chemical mechanisms, methodologies, or specific examples within a particular domain. The judge model is either the Gemini 3.1 Pro or the DeepSeek-R1. The DeepSeek-R1 serves as the judge during the adaptive training process, whereas the Gemini 3.1 Pro is used for the final ChemJudge benchmark evaluation. During the adaptive training phase, the judge provides a composite score based on scientific accuracy, hallucination, and structured output. For the benchmark evaluation, however, to mitigate biases that LLMs may have toward different output formats, we exclusively assess the hallucination score. This is quantified by the number of hallucinations observed: starting from a total of 10 points per question, one point is deducted for each instance of hallucination. If a response is deemed irrelevant by the judge model, all points for that question are deducted. Based on this scoring, the hallucination rate is calculated from the accumulated deductions over the ChemJudge question set.

To rigorously validate the reliability of the automated ChemJudge metrics a systematic human verification process was implemented. A panel of domain experts comprising doctoral researchers from the fields of inorganic organic physical and analytical chemistry conducted a blind review of the evaluation results. The experts independently assessed the presence of hallucinations and verified the validity of the deductions made by the judge model. Crucially to ensure the absolute accuracy of the benchmark a strict priority protocol was established whereby in any event of a discrepancy between the score assigned by the Gemini 3.1 Pro judge and the assessment of the human expert the human expert judgment was treated as the immutable ground truth. Consequently the final hallucination rates reported in this study reflect expert verified data ensuring that the evaluation is not compromised by potential errors in the judge model.

\subsection*{OmniChem-DeepResearch}
OmniChem-DeepResearch is an AI assistant designed to provide comprehensive, intelligent support for chemistry researchers across core research activities, including literature review, knowledge acquisition, and professional report generation. The implementation of its core functionality relies on invoking LLM APIs. This design endows DeepResearch with significant flexibility and scalability, enabling seamless integration with any LLM accessible via an API, including both commercial models and open-source models deployed in a local environment. The key functional modules of DeepResearch work in synergy to deliver robust research support. The literature retrieval and content acquisition module utilizes the Web of Science\texttrademark{} Starter API provided by Clarivate. Users can perform efficient and precise literature searches by inputting keywords, for which the system returns a list of DOIs for relevant publications. Subsequently, DeepResearch uses these DOIs to directly access the original publication webpages (typically on official journal websites) and intelligently extracts the core content by parsing HTML snapshots of the pages. Crucially, the completeness of the content retrieved through this process depends on the user's institutional subscription privileges for the relevant academic databases. For non-subscribed resources, the system is generally limited to fetching the publicly available abstract. The designed DeepResearch framework is compatible with API calls to LLMs. Therefore, to further enhance the effectiveness of DeepResearch, it is recommended to utilize APIs from larger-scale models.

\subsection*{Deep Research Evaluation}
The evaluation of Deep Research is conducted through a comprehensive assessment framework totaling 100 points, which is structured around four equally weighted dimensions. The first dimension, Information Accuracy, accounts for 25 points and scrutinizes the factual correctness and reliability of the provided content. Another 25 points are allocated to Depth and Logic, which assesses the thoroughness of the analysis and the coherence of its reasoning. The third dimension, Structure and Expression, also worth 25 points, evaluates the organizational quality and clarity of the presentation. Finally, Innovation and Practicality are judged for the remaining 25 points, measuring the novelty of the insights and their real-world applicability. The quantitative scoring table is provided in the Supplementary Information.

\subsection*{OmniChem-QA Expert}
OmniChem integrates a dynamically constructed module, QA-Expert, which is architecturally based on an advanced RAG framework and powered by a continuously evolving, domain-specific knowledge graph as its core knowledge base. Within this system, designated LLMs are tasked with the in-depth analysis of extensive domain literature to automatically identify and extract key concepts, critical entities, and their intricate semantic relationships (e.g., clustering, causality, and hierarchical dependencies). This process facilitates the construction of a highly structured content index, culminating in a dynamically updated and progressively enriched knowledge graph for the chemistry domain. All LLM invocations for the construction, maintenance, and querying of the knowledge graph are managed through standardized API calls. For text embedding representation, the system employs the mxbai-embed-large-v1 model to ensure the precise capture of semantic information. Furthermore, to enhance user experience, an interactive visualization interface developed with the Gradio framework has been integrated. This interface significantly improves the convenience, intuitiveness, and efficiency of knowledge exploration, information retrieval, and complex result analysis for the user.

\subsection*{Absorption Wavelength Calculation}
All quantum chemical calculations were performed using the Gaussian 6 software package. Ground-state geometries were fully optimized using the B3LYP functional with the 6-311G(d) basis set. Vibrational frequency calculations at the same level of theory were conducted to confirm all stationary points as local minima. Subsequently, absorption wavelengths were computed using time-dependent density functional theory (TD-DFT). The key TD-DFT transition table and orbital visualizations are provided in the Supplementary Information.

\subsection*{Aqueous Solubility Calculation}
\section*{Acknowledgments}
The authors would like to acknowledge the financial support from the National Natural Science Foundation of China (No. 22372025, 22272017), the Excellent Youth Fund of Liaoning Province (2024JH3/10200005), the Fundamental Research Funds for the Central Universities (No. DUT25Z2722, DUT22LAB607).

\section*{Author contributions}
Conceptualization: R.L., W.Y., S.T., Z.W.\\
Methodology: B.B., R.L., L.Y., Z.W., Y.M., X.L.\\
Investigation: R.L., W.Y., S.T., Z.W.\\
Visualization: R.L., Y.M.\\
Funding acquisition: S.T., W.Y.\\
Project administration: R.L, W.Y., S.T.\\
Supervision: W.Y., S.T.\\
Writing -- original draft: R.L., W.Y., S.T., B.B., H.Y., Z.W.\\
Writing -- review \& editing: R.L., W.Y., S.T.\\

\section*{Data and materials availability}
The open-source code for the G-Frame framework, along with all benchmark datasets used for evaluation, is publicly available on GitHub under the MIT license (\url{https://github.com/Billy-Liu-DUT/G-Frame}). The final 7B parameter chemical expert model, OmniChem-7B-v1, and the corresponding OmniChem dataset used for fine-tuning are available at the Hugging Face Hub (\url{https://huggingface.co/Billy-Liu-DUT/OmniChem-7B-v1} and \url{https://huggingface.co/datasets/Billy-Liu-DUT/OmniChem} respectively) under a CC BY-NC-SA 4.0 license. To facilitate reproducibility, the machine-readable XYZ coordinates for all DFT optimized structures, supplementary quantitative tables in CSV format, and original raw calculation reports from ACDLabs software have been deposited in Zenodo (DOI: 10.5281/zenodo.18446987). All other data supporting the findings of this study are available within the main text or the supplementary materials.

\section*{Competing interests}
Authors declare that they have no competing interests.

\clearpage
\appendix
\section{Supplementary Information}
This appendix contains the supplementary evidence needed to support the arXiv version of the manuscript: adaptive concurrency data, evaluation tables, chemical design calculations, a concrete hierarchical-game mapping, and the full ablation study.

\section{Mathematical Basis of G-Frame}
The key mathematical definitions from the supplementary theory section are retained here in LaTeX form. Routine explanatory text and repeated symbolic definitions are condensed to keep the arXiv supplement readable.

For an input context $x$, text generation by an LLM with parameters $\Phi$ is modeled as an autoregressive conditional probability process over the output sequence $y=(y_1,y_2,\ldots,y_T)$:
\begin{equation}
P_{\Phi}(y\mid x)=\prod_{t=1}^{T}P_{\Phi}(y_t\mid y_{<t},x).
\end{equation}
The one-step token probability is defined by the softmax output of the last-layer hidden state $h_t$ and embedding matrix $E$:
\begin{equation}
P_{\Phi}(y_t\mid y_{<t},x)=\operatorname{Softmax}(h_t E^{\mathsf{T}}).
\end{equation}
Given an ideal low-entropy truth distribution $P_{\mathrm{true}}(y\mid x)$, hallucination is formalized as divergence between the generated and true distributions:
\begin{equation}
D_{\mathrm{KL}}\!\left(P_{\mathrm{true}}\Vert P_{\Phi}\right)
=\sum_y P_{\mathrm{true}}(y\mid x)\log\frac{P_{\mathrm{true}}(y\mid x)}{P_{\Phi}(y\mid x)}>\epsilon.
\end{equation}
The team-game mechanism introduces intermediate latent variables $z=\{z_1,z_2,\ldots,z_n\}$ and role constraints $r=\{r_1,r_2,\ldots,r_n\}$, decomposing the long generation path into short constrained transitions:
\begin{equation}
P_{\mathrm{team}}(y\mid x)
=\int P(y\mid z_n,r_n)\prod_{i=1}^{n}P(z_i\mid z_{i-1},r_i)\,dz.
\end{equation}
A verifier agent assigns a confidence score $S(z_i)\in[0,1]$ to intermediate states. This acts as importance weighting on the original model distribution:
\begin{equation}
P_{\mathrm{new}}(y\mid x)\propto P_{\Phi}(y\mid x)\prod_i S(z_i),
\qquad
P_{\mathrm{new}}(y\mid x)\propto P_{\Phi}(y\mid x)S(y).
\end{equation}
The expected result is a sharpened distribution closer to the factual target:
\begin{equation}
\min_{r_i}D_{\mathrm{KL}}\!\left(P_{\mathrm{true}}\Vert P_{\mathrm{team}}\right)
\ll D_{\mathrm{KL}}\!\left(P_{\mathrm{true}}\Vert P_{\Phi}\right).
\end{equation}
The Bayesian macro-control problem is expressed as a partially observable decision process $\langle S,A,T,R,\Omega\rangle$, with hidden state $\theta=(\theta_{\mathrm{task}},\theta_{\mathrm{sys}})$ and decision $d\in A$. The global utility balances generation quality, resource cost, and a large penalty for system failure:
\begin{equation}
U(d,\theta)=\operatorname{QualityGain}(d,\theta)
-\beta\,\operatorname{ResourceCost}(d,\theta)-\lambda\,\mathbb{I}_{\mathrm{Crash}},
\qquad \lambda\rightarrow\infty.
\end{equation}
Because $\theta$ is not directly observed, the decisional agent maintains a belief state $b_t(\theta)=P(\theta\mid \Gamma_{1:t})$. After receiving observation $\Gamma_t$, belief is updated by Bayes' rule:
\begin{equation}
b_t(\theta)\propto P(\Gamma_t\mid \theta,d_{t-1})\,b_{t-1}(\theta).
\end{equation}
The posterior decision is then selected by maximizing expected utility under the updated belief:
\begin{equation}
d_t^{*}=\arg\max_{d\in D}\mathbb{E}_{\theta\sim b_t(\theta)}[U(d,\theta)]
=\arg\max_{d\in D}\int U(d,\theta)b_t(\theta)\,d\theta.
\end{equation}

\section{Adaptive Concurrency Control}
To address the challenge of efficiently utilizing large language model (LLM) inference services for complex and dynamic tasks, we present an in-depth analysis of a multi-agent framework predicated on asynchronous concurrency and adaptive control. This framework decouples system functionalities into three core hierarchical components---a computational engine layer, a task execution layer, and a central decision agent---to construct a robust, efficient, and self-optimizing closed-loop system. This section provides a detailed, progressive examination of the design philosophy and core implementation of each component, elucidates the data and control flow relationships between them, and ultimately reveals how this architecture systematically resolves performance bottlenecks and service interruptions that LLM services may encounter when facing heterogeneous workloads.

These three components collectively form an elegant, decoupled, closed-loop feedback control system. The complete control flow begins with the initialization by the decision agent, which concurrently starts the computational engine and the worker agents. The worker agents then commence sending inference requests according to the current concurrency limit. Simultaneously, the computational engine and worker agents report server load and task progress, respectively, to the decision agent in real-time and asynchronously through their monitoring modules. Upon fusing these two information streams, the decision agent achieves full situational awareness of the system's state. Based on its internal adaptive logic, it determines whether to increase concurrency to enhance efficiency or decrease it to prevent service overload. Finally, the decision agent applies its new concurrency decision directly to the dynamic semaphore used by the worker agents, thereby altering their subsequent request behavior. This change in behavior immediately impacts the load state of the computational engine, and this new state is in turn captured by the monitoring system and fed back to the decision agent, thus creating a continuous, self-optimizing loop. In summary, this architecture, through clear separation of responsibilities and data-driven real-time feedback, achieves intelligent regulation of critical system parameters. It ensures that the LLM inference service can consistently maintain an optimal balance between high efficiency and stability when faced with complex and dynamic scientific computing tasks, offering a valuable and generalizable engineering paradigm for resolving performance optimization challenges in the practical application of large models.

The robust operation of this framework relies on a decoupled, asynchronous multi-queue communication architecture built upon the asyncio library. The design adheres to a Multi-Producer, Single-Consumer event-driven pattern, ensuring efficient, reliable, and non-blocking information exchange between components, which is the technical cornerstone for achieving real-time adaptive control.

In summary, this communication mechanism achieves functional decoupling through asynchronous queues, ensures immediate responsiveness via concurrent listening, and enables precise fusion of heterogeneous data through a timestamp synchronization protocol. This suite of design features ensures that the decision agent continuously and reliably receives high-quality feedback on the system's state, providing a solid data foundation for its subsequent complex decision-making process based on Bayesian games.

Here, we present statistical graphs showing the KV Cache usage and the status of pending tasks for the vLLM engine under different processing modes: synchronous processing, fixed asynchronous concurrency levels of 50, 100, 200, and 500, and adaptive optimization using G-Frame.

The test involved approximately 8,600 tasks. Here, we present the total number of tokens required to complete them.

The adaptive optimization enabled by G-Frame is key to enhancing processing efficiency. When handling identical tasks, it consistently maintains the vLLM engine's KV Cache usage at an optimal 80-90\% level. This strategy achieves a critical balance: it maximizes the utilization of GPU memory while preserving a safe headroom in the KV Cache. This headroom is essential for preventing a large queue of pending processes from causing timeout errors, particularly when faced with more computationally demanding tasks.

\begin{figure}[!htbp]
\centering
\includegraphics[width=\linewidth]{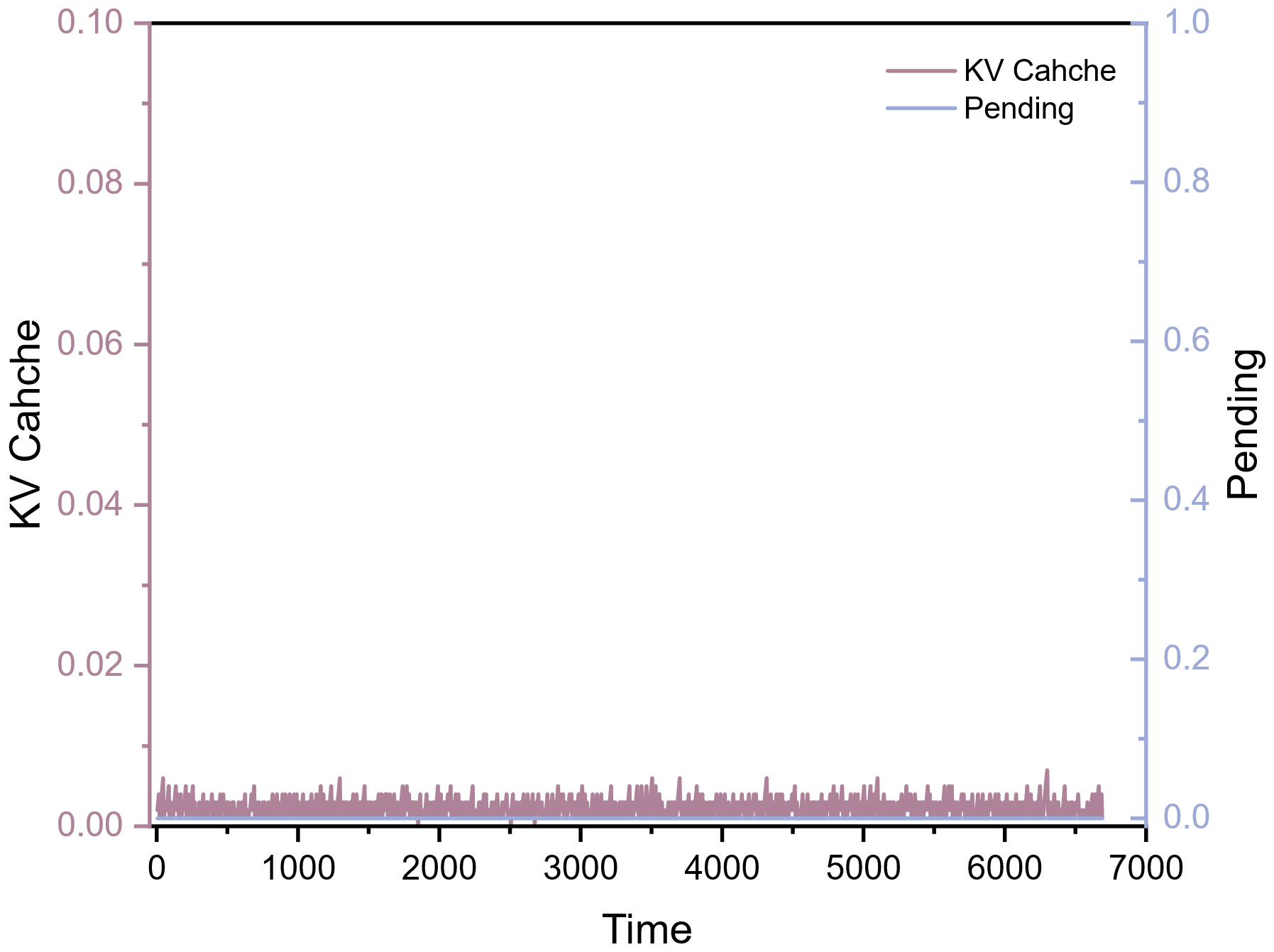}
\caption{KV Cache usage and pending process statistics in synchronous mode.}
\label{fig:figs1}
\end{figure}

\begin{figure}[!htbp]
\centering
\includegraphics[width=\linewidth]{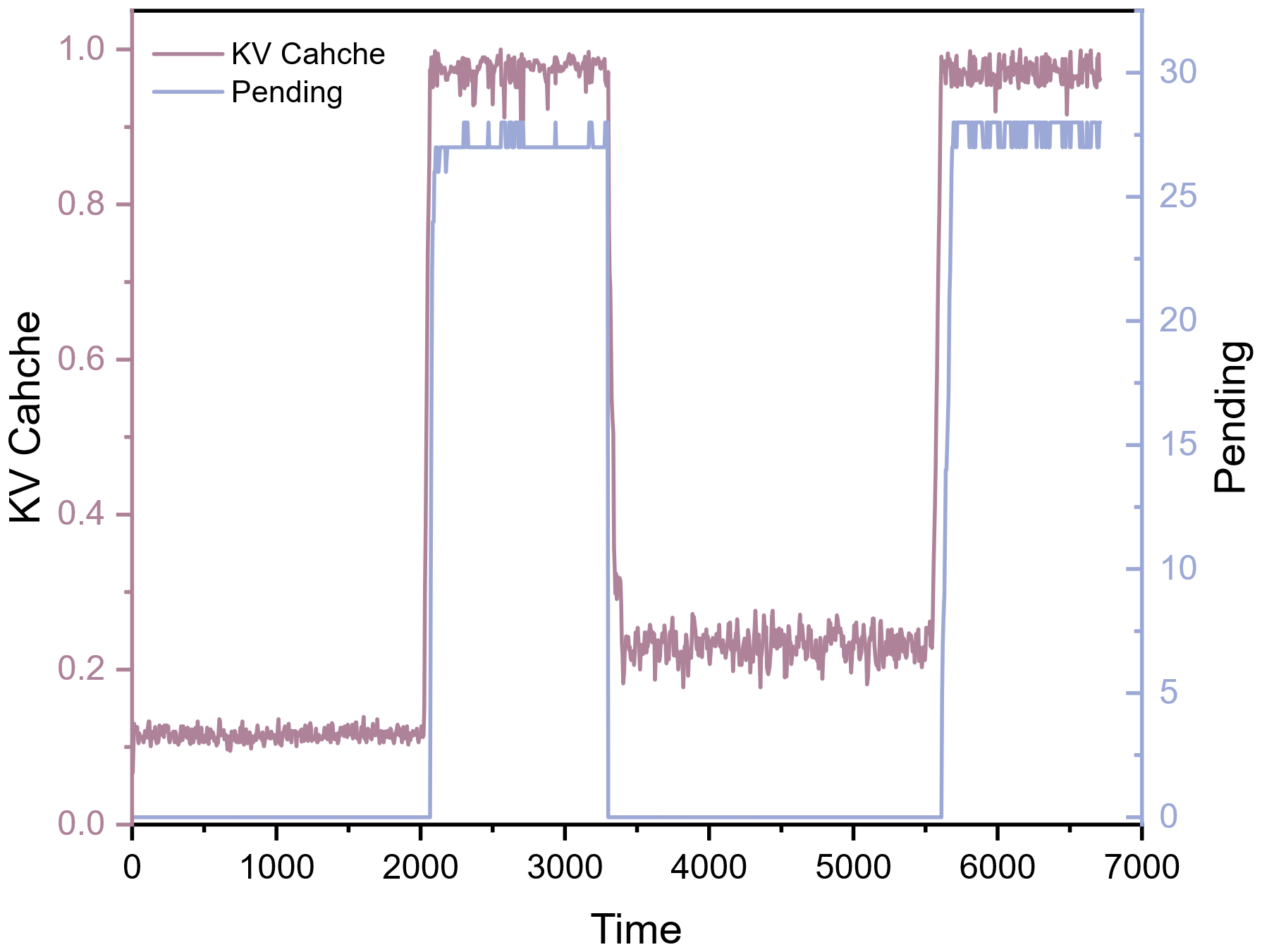}
\caption{KV Cache usage and pending process statistics at a fixed concurrency level of 50.}
\label{fig:figs2}
\end{figure}

\begin{figure}[!htbp]
\centering
\includegraphics[width=\linewidth]{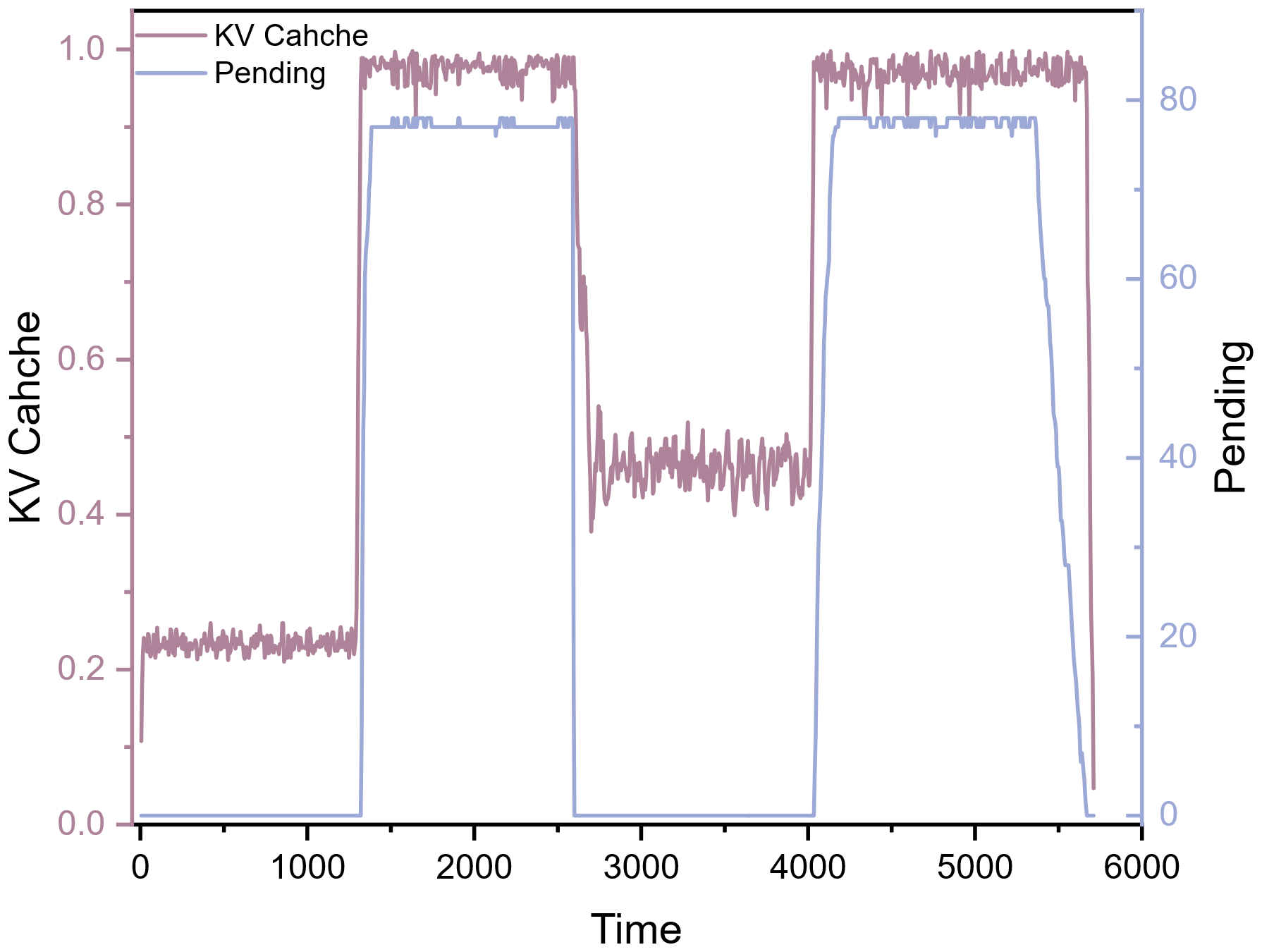}
\caption{KV Cache usage and pending process statistics at a fixed concurrency level of 100.}
\label{fig:figs3}
\end{figure}

\begin{figure}[!htbp]
\centering
\includegraphics[width=\linewidth]{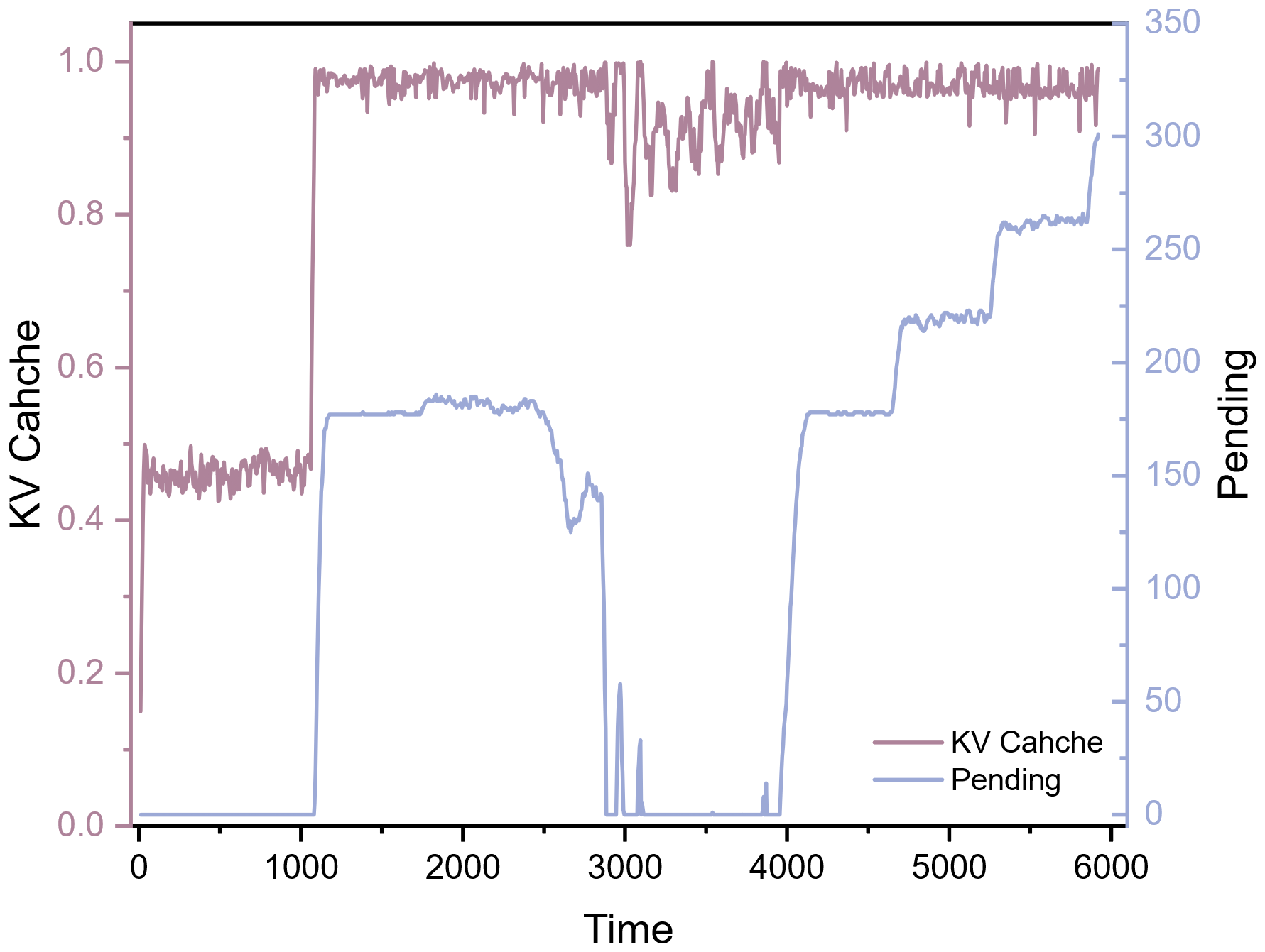}
\caption{KV Cache usage and pending process statistics at a fixed concurrency level of 200.}
\label{fig:figs4}
\end{figure}

\begin{figure}[!htbp]
\centering
\includegraphics[width=\linewidth]{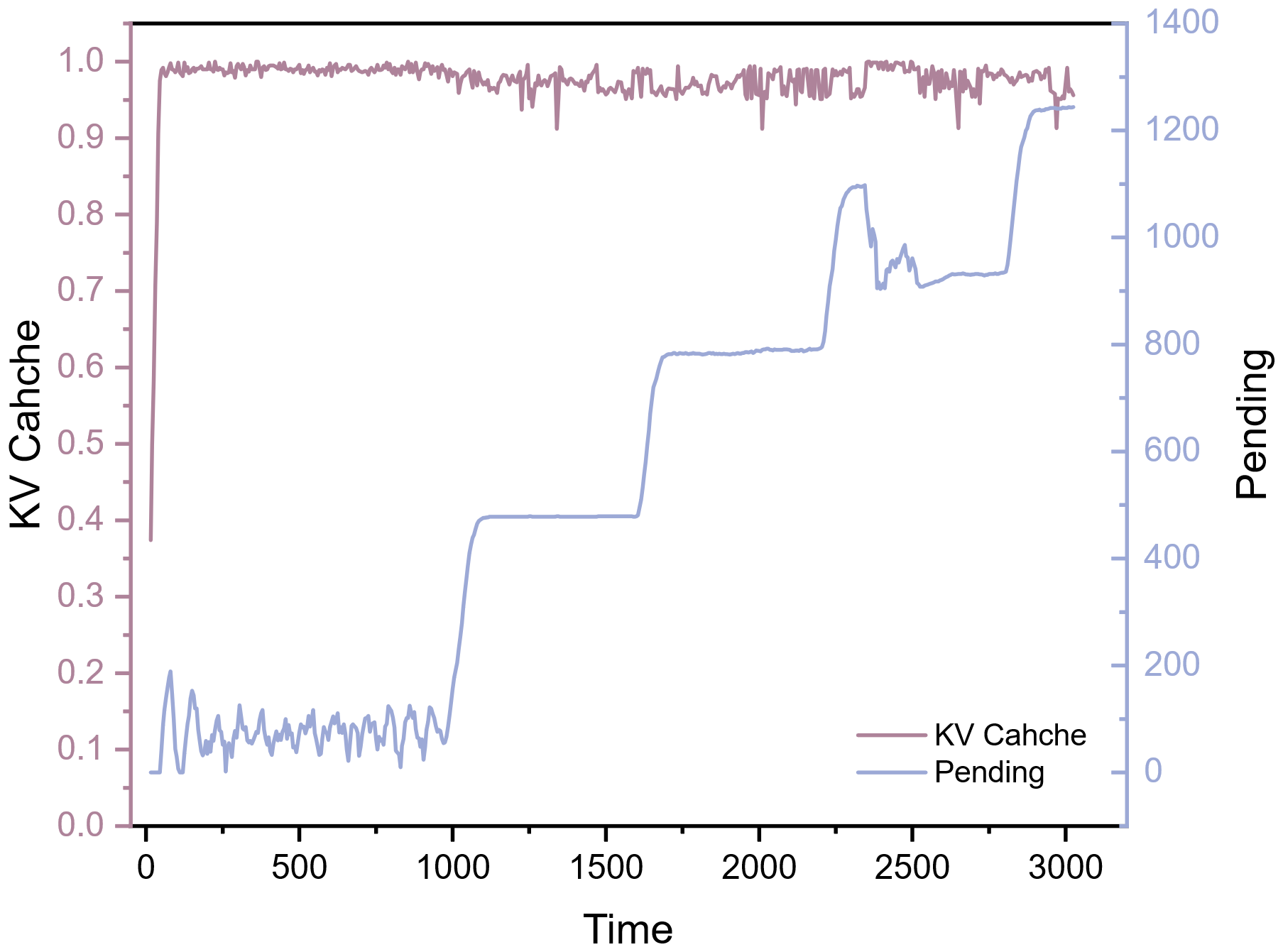}
\caption{KV Cache usage and pending process statistics at a fixed concurrency level of 500.}
\label{fig:figs5}
\end{figure}

\begin{figure}[!htbp]
\centering
\includegraphics[width=\linewidth]{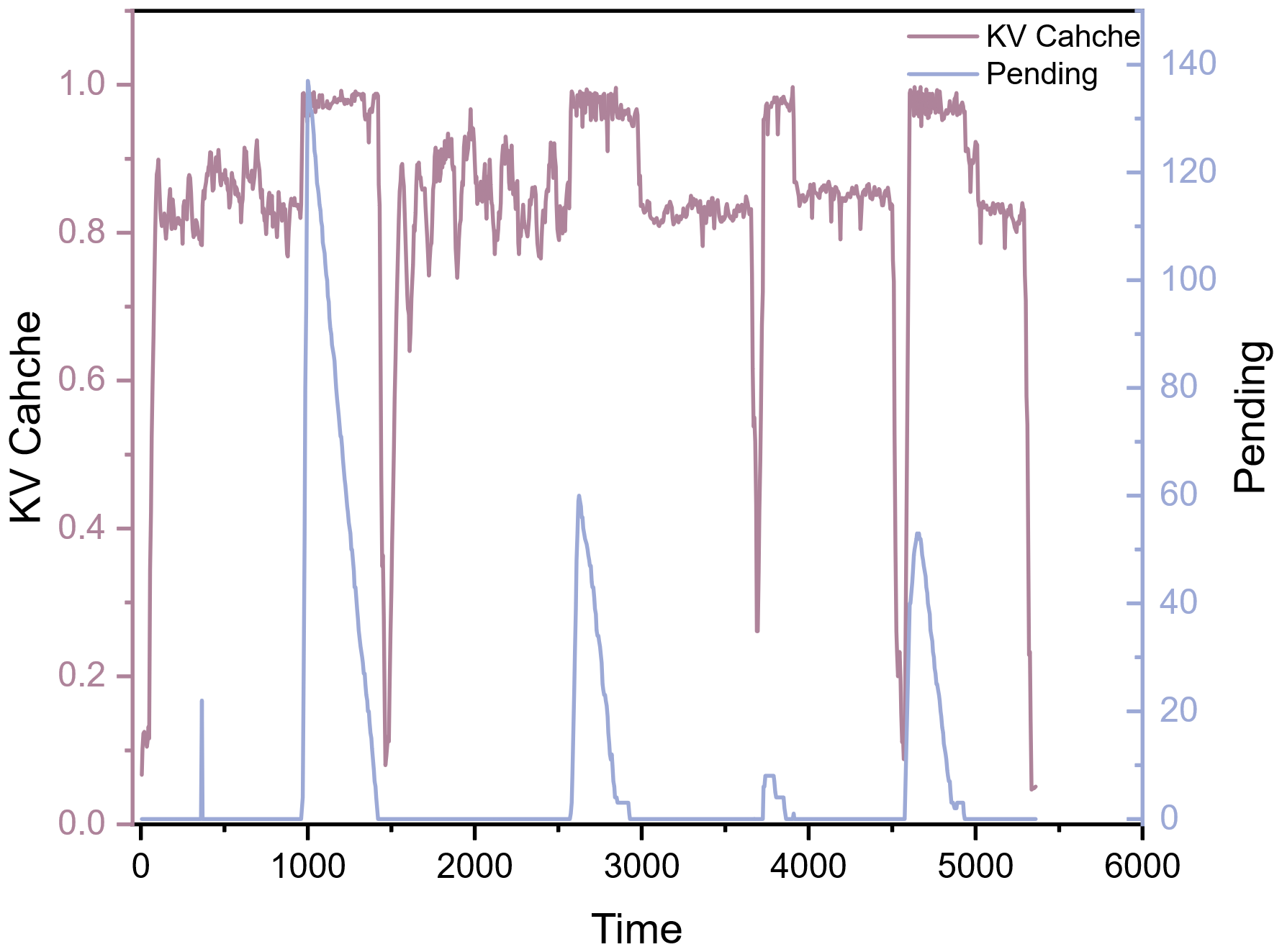}
\caption{KV Cache usage and pending process statistics under G-Frame adaptive concurrency control.}
\label{fig:figs6}
\end{figure}

\begin{figure}[!htbp]
\centering
\includegraphics[width=\linewidth]{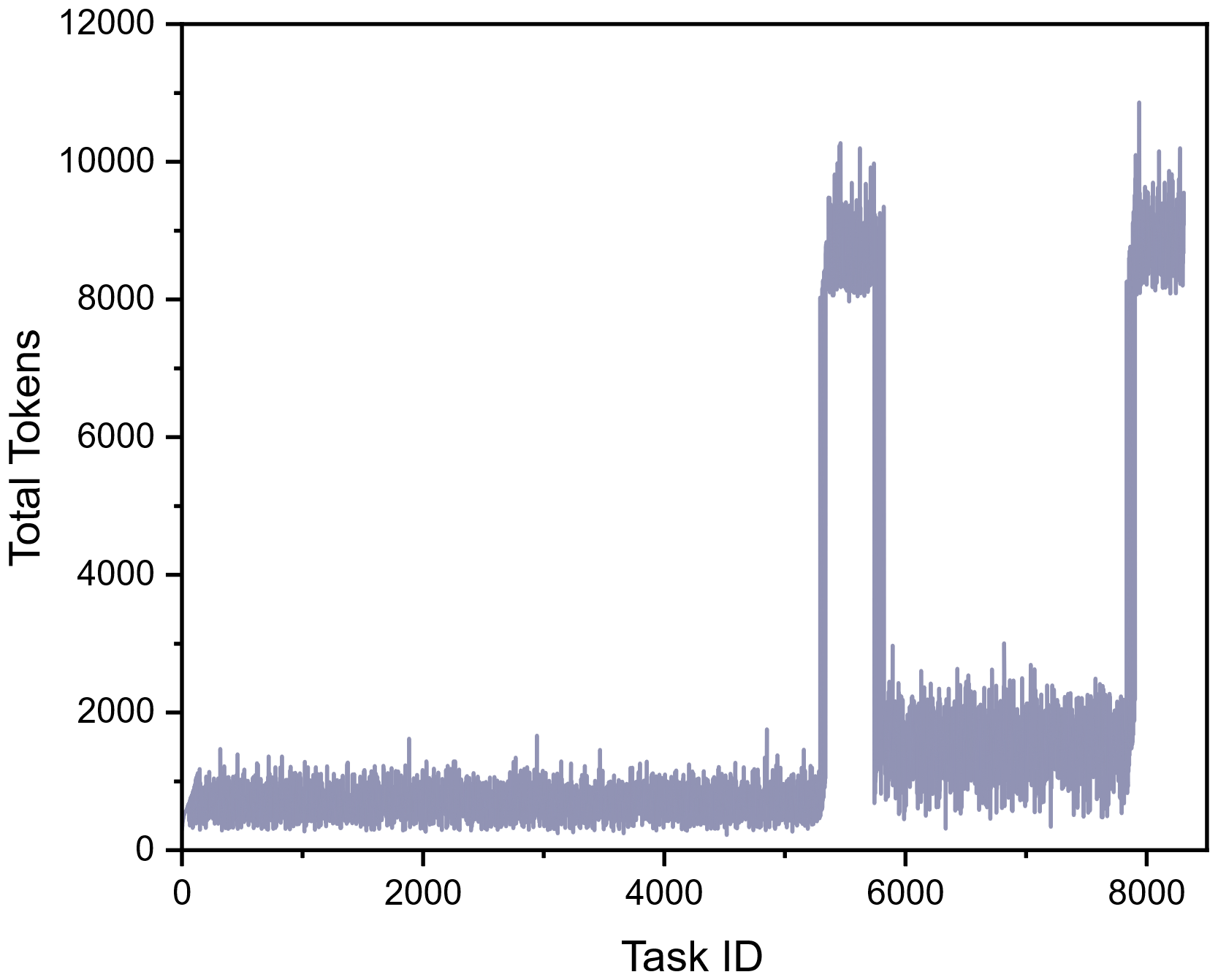}
\caption{Token consumption statistics for the test tasks.}
\label{fig:figs7}
\end{figure}

\section{Evaluation Protocols and Quantitative Tables}
\textbf{ThChem}\textbf{ Benchmark}

ThChem is a benchmark meticulously constructed to accurately assess the comprehensive capabilities of large language models in the domain of chemistry. Its content extensively covers core branches of the discipline, including inorganic, organic, analytical, and physical chemistry. The benchmark comprises two iterative versions: ThChem1.0 and ThChem2.0. ThChem1.0 consists of 253 high-quality, multiple-choice questions. These questions not only assess fundamental knowledge---such as chemical equation calculations, substance structure inference, and experimental and characterization techniques---but also extend to advanced topics, including cutting-edge concepts in new material design and their potential applications. All questions are derived from real research cases and complex problems found in high-impact, peer-reviewed academic journals. They have been rigorously cross-validated by multiple senior chemistry experts to ensure scientific accuracy, clarity, and the uniqueness and correctness of the answers. Building upon this foundation, ThChem2.0 introduces a targeted modification. For approximately half of the questions randomly selected from ThChem1.0, the correct answer option was deliberately removed, rendering the standard answer as "null" or non-existent among the choices. This design is intended to evaluate the model's core abilities at a deeper level---specifically, its capacity for complex mechanism inference, multi-step logical reasoning, and the integrated application of chemical knowledge---rather than simply its ability to identify the correct answer from a given set of options.

\textbf{ChemJudge}\textbf{ Benchmark}

This benchmark utilizes the "LLM as a Judge" paradigm. It is composed of 471 in-depth chemistry questions, each designed to probe fundamental theories, chemical mechanisms, methodologies, or specific examples within a particular domain. The judge model is either the Gemini 3.1 Pro or the DeepSeek-R1. The DeepSeek-R1 serves as the judge during the adaptive training process, whereas the Gemini 3.1 Pro is used for the final ChemJudge benchmark evaluation. During the adaptive training phase, the judge provides a composite score based on scientific accuracy, hallucination, and structured output. For the benchmark evaluation, however, to mitigate biases that LLMs may have toward different output formats, we exclusively assess the hallucination score. This is quantified by the number of hallucinations observed: starting from a total of 10 points per question, one point is deducted for each instance of hallucination. If a response is deemed irrelevant by the judge model, all points for that question are deducted. Based on this scoring, the hallucination rate is calculated from the accumulated deductions over the ChemJudge question set.

To rigorously validate the reliability of the automated ChemJudge metrics a systematic human verification process was implemented. A panel of domain experts comprising doctoral researchers from the fields of inorganic organic physical and analytical chemistry conducted a blind review of the evaluation results. The experts independently assessed the presence of hallucinations and verified the validity of the deductions made by the judge model. Crucially to ensure the absolute accuracy of the benchmark a strict priority protocol was established whereby in any event of a discrepancy between the score assigned by the Gemini 3.1 Pro judge and the assessment of the human expert the human expert judgment was treated as the immutable ground truth. Consequently the final hallucination rates reported in this study reflect expert verified data ensuring that the evaluation is not compromised by potential errors in the judge model.

\textbf{Deep Research Evaluation}

The evaluation of Deep Research is conducted through a comprehensive assessment framework totaling 100 points, which is structured around four equally weighted dimensions. The first dimension, Information Accuracy, accounts for 25 points and scrutinizes the factual correctness and reliability of the provided content. Another 25 points are allocated to Depth and Logic, which assesses the thoroughness of the analysis and the coherence of its reasoning. The third dimension, Structure and Expression, also worth 25 points, evaluates the organizational quality and clarity of the presentation. Finally, Innovation and Practicality are judged for the remaining 25 points, measuring the novelty of the insights and their real-world applicability. A detailed breakdown of the scoring for each section is provided in the Supplementary Information.

\begin{table}[!htbp]
\centering
\tiny
\setlength{\tabcolsep}{1pt}
\renewcommand{\arraystretch}{1.15}
\caption{Deep Research evaluation scores across judge models. Scores are reported on four 25-point dimensions and summed to an aggregate 100-point score.}
\label{tab:deep-research-scores}
\begin{tabular}{@{}>{\raggedright\arraybackslash}p{0.13\linewidth}>{\raggedright\arraybackslash}p{0.13\linewidth}>{\raggedright\arraybackslash}p{0.11\linewidth}>{\raggedright\arraybackslash}p{0.11\linewidth}>{\raggedright\arraybackslash}p{0.15\linewidth}>{\raggedright\arraybackslash}p{0.16\linewidth}>{\raggedright\arraybackslash}p{0.09\linewidth}@{}}
\toprule
\textbf{Generated report} & \textbf{Judge model} & \textbf{\shortstack{Information\\Accuracy}} & \textbf{\shortstack{Depth\\and Logic}} & \textbf{\shortstack{Structure\\and Expression}} & \textbf{\shortstack{Innovation\\and Practicality}} & \textbf{\shortstack{Aggregate\\Score}} \\
\midrule
DeepResearch-OC & ChatGPT-o3 & 22 & 21 & 20 & 18 & 81 \\
DeepResearch-OC & Gemini2.5Pro & 25 & 24 & 25 & 21 & 95 \\
DeepResearch-OC & Deepseek-R1 & 18 & 22 & 21 & 20 & 81 \\
Gemini2.5Pro & ChatGPT-o3 & 23 & 24 & 23 & 22 & 92 \\
Gemini2.5Pro & Gemini2.5Pro & 24 & 25 & 25 & 25 & 99 \\
Gemini2.5Pro & Deepseek-R1 & 24 & 25 & 23 & 25 & 97 \\
ChatGPT-o3 & ChatGPT-o3 & 23 & 24 & 22 & 21 & 90 \\
ChatGPT-o3 & Gemini2.5Pro & 25 & 24 & 25 & 24 & 98 \\
ChatGPT-o3 & Deepseek-R1 & 25 & 25 & 25 & 25 & 100 \\
\bottomrule
\end{tabular}
\end{table}

\begin{figure}[!htbp]
\centering
\includegraphics[width=\linewidth]{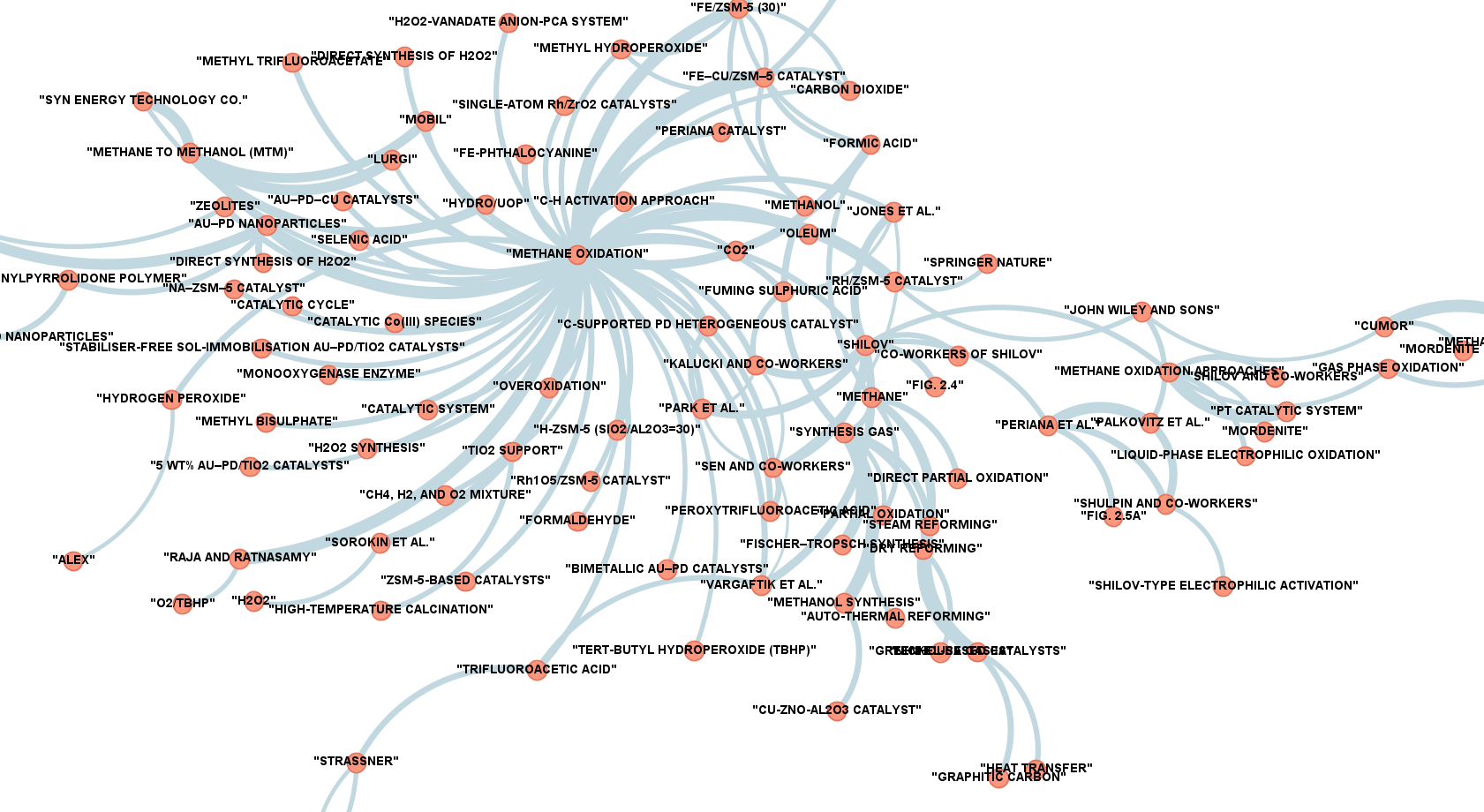}
\caption{Original knowledge graph generated by the QA Expert module for the domain-specific chemistry example.}
\label{fig:figs12}
\end{figure}

\section{Chemical Design Calculations}
Here, we present a design case study in which OmniChem was tasked with creating BODIPY derivatives with targeted red-light absorption properties.

Design of BODIPY Molecules with Enhanced Water Solubility

Here, we demonstrate an example of using OmniChem to design BODIPY molecules with improved water solubility.

We present the atomic coordinates of the optimized geometry. All calculations were performed using the B3LYP functional, augmented with the GD3BJ empirical dispersion correction, and the 6-311G(d) basis set. The ground-state geometry was optimized in a simulated acetonitrile solvent environment, utilizing the Integral Equation Formalism variant of the Polarizable Continuum Model (IEFPCM). A subsequent vibrational frequency analysis was performed to confirm the optimized structure as a true energy minimum. Finally, single-point TD-DFT calculations were carried out on the optimized ground-state geometry to obtain the vertical transition energies for the first thirty excited states.

The key TD-DFT transition data and orbital visualizations are summarized below.

\begin{table}[!htbp]
\centering
\scriptsize
\setlength{\tabcolsep}{3pt}
\renewcommand{\arraystretch}{1.15}
\caption{TD-DFT transition energies and oscillator strengths for the red-absorbing BODIPY design.}
\label{tab:absorption-transition}
\begin{tabular}{@{}>{\raggedright\arraybackslash}p{0.16\linewidth}>{\raggedright\arraybackslash}p{0.20\linewidth}>{\raggedright\arraybackslash}p{0.20\linewidth}>{\raggedright\arraybackslash}p{0.12\linewidth}>{\raggedright\arraybackslash}p{0.18\linewidth}@{}}
\toprule
\textbf{State} & \textbf{Energy (eV/nm)} & \textbf{Composition} & \textbf{f} & \textbf{Character} \\
\midrule
S1 & 1.47/845 & H+1L & 0.2695 & CT \\
S2 & 1.75/709 & HL & 0.8971 & CT \\
\bottomrule
\end{tabular}
\end{table}

\begin{figure}[!htbp]
\centering
\includegraphics[width=\linewidth]{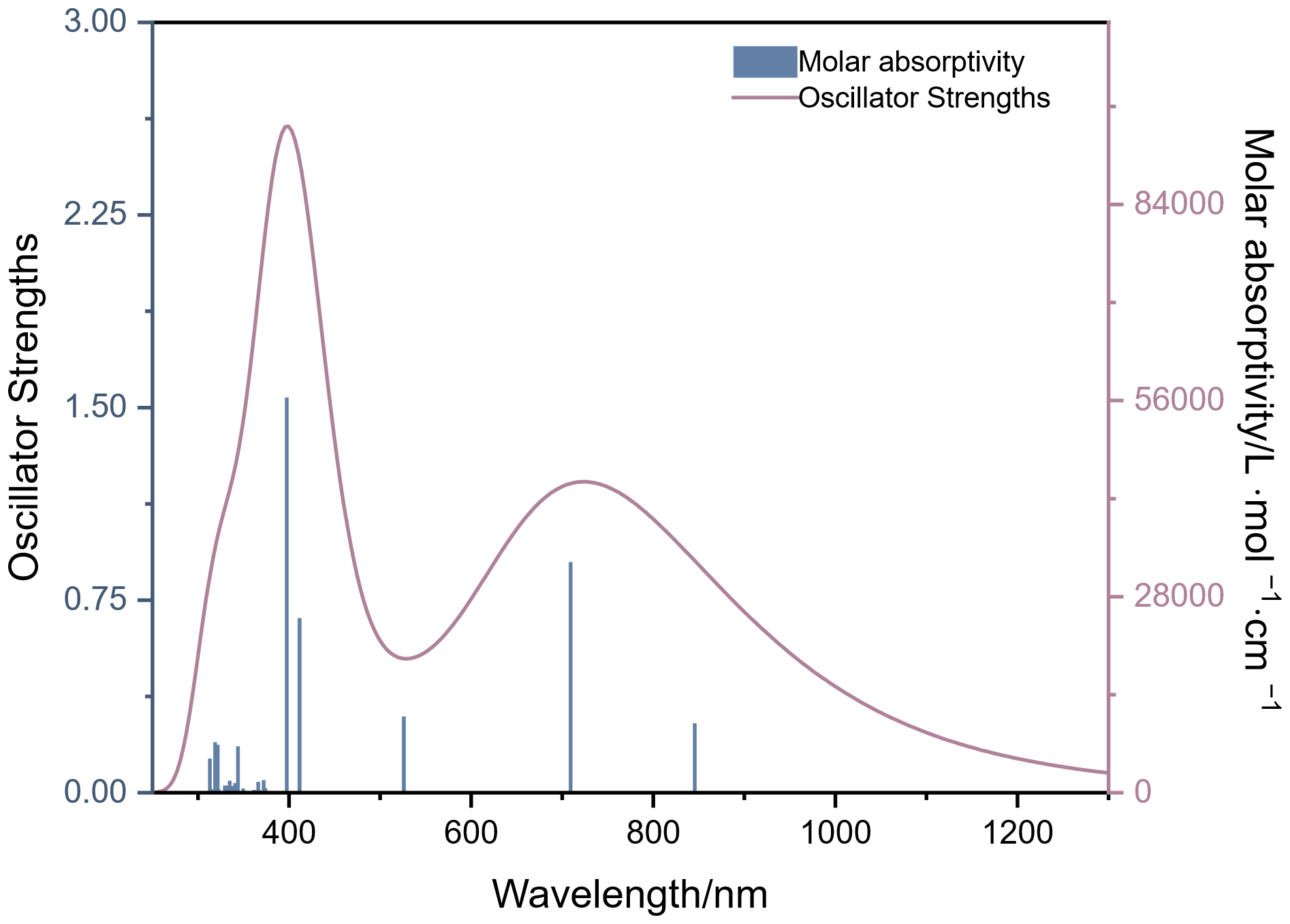}
\caption{Simulated UV-Vis-NIR absorption spectrum and corresponding oscillator strengths for the red-absorbing BODIPY design.}
\label{fig:figs8}
\end{figure}

\begin{figure}[!htbp]
\centering
\includegraphics[width=\linewidth]{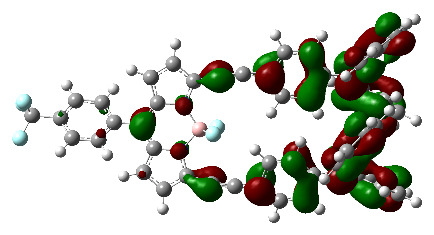}
\caption{Molecular orbital isosurface visualization for HOMO.}
\label{fig:figs9}
\end{figure}

\begin{figure}[!htbp]
\centering
\includegraphics[width=\linewidth]{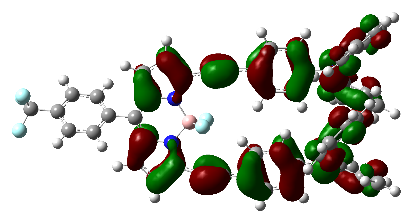}
\caption{Molecular orbital isosurface visualization for HOMO+1.}
\label{fig:figs10}
\end{figure}

\begin{figure}[!htbp]
\centering
\includegraphics[width=\linewidth]{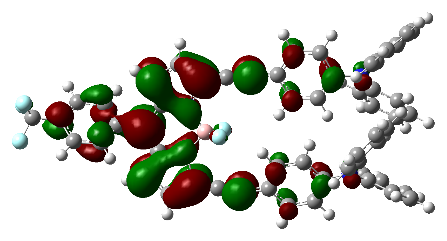}
\caption{Molecular orbital isosurface visualization for LUMO.}
\label{fig:figs11}
\end{figure}

\section{Concrete Running Example of the Hierarchical Game}
To enhance the clarity of the mathematical symbols presented in Fig. 2a and Algorithm 2, we provide a concrete, step-by-step running example. This section illustrates how the abstract symbols map to specific operational data flows within the micro-level "Team Game," executed via the multi-agent prompts (Process i through Process v) described in the implementation details.

In this phase, the system executes specific text generation tasks under the strategy constraints (such as concurrency and sampling depth) determined by the macro-level decision layer. The objective is to optimize the quality of the generated content through multi-agent collaboration.

\textbf{Scenario:} Synthesis of QA pairs from a specific academic paper.

\textbf{20}\textbf{.2 }\textbf{Macro-Level Execution: The Bayesian Game (Resource Scheduling)}

\textbf{Scenario:} The system is processing the complex chemical literature batch described in Section 20.1. As the verification tasks regarding the TBADT reaction mechanism intensify, the computational graph expands, causing a rapid spike in memory demand.

\begin{table}[!htbp]
\centering
\scriptsize
\setlength{\tabcolsep}{3pt}
\renewcommand{\arraystretch}{1.15}
\caption{Mapping between micro-level team-game concepts and their runtime implementation in the QA-generation example.}
\label{tab:team-game-mapping}
\begin{tabular}{@{}>{\raggedright\arraybackslash}p{0.24\linewidth}>{\raggedright\arraybackslash}p{0.36\linewidth}>{\raggedright\arraybackslash}p{0.30\linewidth}@{}}
\toprule
\textbf{Definition} & \textbf{Component in user prompts} & \textbf{Runtime value example} \\
\midrule
Input Data & \{Article\} & The text on TBADT photocatalysis. \\
Role Constraints & Prompt System Instructions & "You are an AI assistant acting as a meticulous Reviewer..." (Process iii) \\
Initial Sample & Process ii Output & The draft answer containing the yield error (50\%). \\
Verification Function & Process iii Logic & The critique identifying the discrepancy between 50\% vs 92\%. \\
Interaction Topology & The Workflow Graph & The cyclic flow: Teacher (i) Student (ii) Teacher (iii) Student (iv). \\
Final Optimized Output & Process v Output & The final, verified answer synthesized by the Judger. \\
\bottomrule
\end{tabular}
\end{table}

\begin{table}[!htbp]
\centering
\scriptsize
\setlength{\tabcolsep}{3pt}
\renewcommand{\arraystretch}{1.15}
\caption{Mapping between macro-level Bayesian-game concepts and their runtime implementation in adaptive concurrency control.}
\label{tab:bayesian-game-mapping}
\begin{tabular}{@{}>{\raggedright\arraybackslash}p{0.28\linewidth}>{\raggedright\arraybackslash}p{0.36\linewidth}>{\raggedright\arraybackslash}p{0.26\linewidth}@{}}
\toprule
\textbf{Theoretical definition} & \textbf{Component in adaptive-concurrency prompt} & \textbf{Runtime value example} \\
\midrule
Prior Strategy (Alg 2, Line 8) & Context History (Previous Action) & qa\_synth --concurrency 16 \\
Feedback/Observation (Alg 2, Line 10) & [Current Conditions] JSON Input & \{"gpu\_vram\_usage": "92\%", ...\} \\
Utility Function (Alg 2, Line 13) & Optimization Objectives in Prompt & "Achieve optimal balance... and system stability" \\
Implicit Bayesian Inference & Reasoning Requirement in Prompt & Logic derived: "High VRAM risk requires throttling." \\
Optimized Decision (Alg 2, Line 13) & Plan\_of\_Action JSON Output & ["qa\_synth --concurrency 8"] \\
Joint Type Space & Task Constraints \& Context & High computational complexity of chemical verification. \\
\bottomrule
\end{tabular}
\end{table}

\section{Ablation Study on Training Stages}
To rigorously isolate and quantify the individual and synergistic contributions of the three core training stages---(i) Domain Pretraining (DPT), (ii) Synthetic QA/CoT Fine-tuning (SFT), and (iii) Judge-guided Adaptive Training (AT)---to hallucination mitigation and factual reasoning, we conducted a full factorial ablation study. By systematically enabling or disabling these three components, we derived 8 distinct model configurations, starting from the unaligned Base LLM to the fully equipped OmniChem architecture. To ensure a fair comparison when ablating the second factor, we maintained an identical training volume by utilizing conventional chemistry question-answering data sourced from the camel-ai/chemistry dataset.

To comprehensively assess the performance of each configuration, all 8 model variants were evaluated across four distinct benchmarks:

ThChem 1.0: To measure the accuracy of domain-specific knowledge retrieval and basic factual reasoning.

ThChem 2.0: To evaluate the model's boundary behavior and its ability to safely refuse unanswerable queries, which serves as a strict metric for hallucination mitigation.

ChemJudge: To assess deep mechanistic reasoning and logical consistency under an LLM-as-a-judge framework.

ChemBench: To provide an independent, third-party validation of zero-shot factual accuracy and generalization capability across a broad spectrum of cheminformatics tasks.

*The hyphen symbol(-) indicates an absence of instruction following where the answer was not generated according to the designated format.

\begin{table}[!htbp]
\centering
\tiny
\setlength{\tabcolsep}{3pt}
\renewcommand{\arraystretch}{1.15}
\caption{Comprehensive ablation study across domain pretraining, synthetic QA/CoT fine-tuning, and adaptive training. A checkmark indicates that the component is included; -- indicates that the component is absent or that the metric was not reported for that configuration.}
\label{tab:ablation}
\begin{tabular}{@{}>{\raggedright\arraybackslash}p{0.06\linewidth}>{\raggedright\arraybackslash}p{0.13\linewidth}>{\raggedright\arraybackslash}p{0.13\linewidth}>{\raggedright\arraybackslash}p{0.13\linewidth}>{\raggedright\arraybackslash}p{0.10\linewidth}>{\raggedright\arraybackslash}p{0.10\linewidth}>{\raggedright\arraybackslash}p{0.10\linewidth}>{\raggedright\arraybackslash}p{0.10\linewidth}@{}}
\toprule
\textbf{Group} & \textbf{\shortstack{Domain\\Pretraining}} & \textbf{\shortstack{Synthetic\\QA/CoT}} & \textbf{\shortstack{Adaptive\\Training}} & \textbf{\shortstack{ThChem\\1.0}} & \textbf{\shortstack{ThChem\\2.0}} & \textbf{ChemJudge} & \textbf{ChemBench} \\
\midrule
1 & -- & \checkmark{} & \checkmark{} & 66.47 & 56.27 & 31.30 & 38.15 \\
2 & \checkmark{} & -- & \checkmark{} & 68.30 & 63.36 & 32.48 & 42.14 \\
3 & \checkmark{} & \checkmark{} & -- & 2.56 & 48.44 & 24.31 & -- \\
4 & -- & -- & \checkmark{} & -- & -- & 57.09 & -- \\
5 & \checkmark{} & -- & -- & -- & 51.05 & 30.17 & -- \\
6 & -- & \checkmark{} & -- & 11.29 & 40.61 & 19.26 & 39.08 \\
7 & -- & -- & -- & 33.82 & 37.15 & 21.68 & 46.47 \\
8 & \checkmark{} & \checkmark{} & \checkmark{} & 79.45 & 62.08 & 4.71 & 49.82 \\
\bottomrule
\end{tabular}
\end{table}

The ablation results demonstrate the independent mechanisms and interdependencies of the three training stages. A complete configuration designated as Group 8 attains the highest evaluation scores on ThChem 1.0 with 79.45 and ChemBench with 49.82. An error rate of 4.71 is recorded on the ChemJudge benchmark indicating minimized hallucination rates. The two stage fine tuning process is maintained unaltered across all experimental groups to ensure consistency and isolate the effects of individual module ablations.

Domain pretraining is implemented to resolve the deficit of specialized professional knowledge. A persistent performance degradation is observed whenever domain pretraining is ablated due to the consequent lack of foundational domain concepts. Group 1 lacking this initial stage yields a score of 66.47 on ThChem 1.0 and 38.15 on ChemBench. The implementation of domain pretraining introduces secondary challenges including repetitive generation and catastrophic forgetting of pretrained weights.

Synthetic question answering fine tuning is utilized to rectify the repetitive generation issue and preserve the acquired professional knowledge. A specific functional duality is observed within this intermediate fine tuning process. The integration of synthetic data with professional knowledge enhances instruction following capabilities and mitigates the forgetting of specialized knowledge when compared to generic tuning data. The full parameter fine tuning process inherently introduces a distinct form of catastrophic forgetting. Configurations retaining domain pretraining but lacking synthetic data fine tuning exhibit pretrained model repetitive generation and literature hallucinations. Group 2 lacking this synthetic fine tuning yields a score of 63.36 on ThChem 2.0 which exceeds the complete model score of 62.08. This elevated score indicates an induced tendency for the model to attempt answering unanswerable queries when instruction tuning is omitted.

Judge guided adaptive training is deployed specifically to resolve the catastrophic forgetting dilemma induced by the preceding fine tuning stage. The ablation of this adaptive training stage consistently triggers catastrophic forgetting across all observed configurations. Group 3 possessing both domain pretraining and synthetic data experiences catastrophic forgetting during the fine tuning stage when adaptive training is removed resulting in a measured ThChem 1.0 score of 2.56. Group 5 possessing domain pretraining but lacking adaptive training experiences catastrophic forgetting earlier during the pretraining phase. A combination of specialized knowledge deficit and catastrophic forgetting is observed in Group 6 where both domain pretraining and adaptive training are ablated. The evaluation of mechanistic reasoning under the ChemJudge framework demonstrates that the integration of domain knowledge and specific reasoning training alters the expression paradigm. This alteration minimizes hallucinations and factual inconsistencies evaluated by the external judge.

Performance degradation pathways are explicitly mapped to distinct module absences to formulate a robust training protocol. The removal of domain pretraining dictates a performance decline due to missing specialized knowledge. The omission of synthetic data fine tuning while retaining pretraining guarantees the emergence of repetitive generation and hallucinations. The exclusion of adaptive training ensures the occurrence of catastrophic forgetting. These empirical observations necessitate a specific adaptive fine tuning strategy. An early stopping mechanism or a reduced learning rate is mandated during the initial training stage to prevent catastrophic forgetting. High quality data must be integrated in the subsequent stage to ensure the acquisition of correct generation formats. This mechanistic separation clarifies that synthetic data slows knowledge degradation while improving instruction following whereas adaptive training rectifies the inevitable performance decline caused by full parameter updates.

\end{document}